\PassOptionsToPackage{prologue,dvipsnames,table}{xcolor}
\documentclass[10pt,twocolumn,letterpaper]{article}

\usepackage[pagenumbers]{cvpr} % To force page numbers, e.g. for an arXiv version

\definecolor{cvprblue}{rgb}{0.21,0.49,0.74}
\usepackage[pagebackref,breaklinks,colorlinks,allcolors=cvprblue]{hyperref}
\usepackage{algorithm}
\usepackage{algorithmic}
\usepackage{amsmath, amssymb}
\usepackage{xcolor}
\usepackage{multirow}
\usepackage[table]{xcolor}
\usepackage{colortbl}
\usepackage{adjustbox}
\usepackage{fancyvrb}
\usepackage{placeins}
\usepackage{etoolbox}
\usepackage{comment}

\def\paperID{6066} % *** Enter the Paper ID here
\def\confName{CVPR}
\def\confYear{2025}

\title{Taming Video Diffusion Prior with Scene-Grounding Guidance \\for 3D Gaussian Splatting from Sparse Inputs
}

\author{
\begin{tabular}[t]{@{}c@{}}
Yingji Zhong$^1$ \quad Zhihao Li$^2$ \quad Dave Zhenyu Chen$^2$ \quad Lanqing Hong$^2$ \quad Dan Xu$^1$
\end{tabular}\\[1ex]
\begin{tabular}[t]{@{}c@{}}
$^1$The Hong Kong University of Science and Technology\quad $^2$Huawei Noah's Ark Lab
\end{tabular}\\[0.5ex]
{\tt\small $\{$yzhongbn,danxu$\}$@cse.ust.hk, $\{$zhihao.li,dave.zhenyuchen,honglanqing$\}$@huawei.com}
}

\begin{document}
\twocolumn[{%
\renewcommand\twocolumn[1][]{#1}%
\maketitle
\begin{center}
    \centering
    \vspace{-12pt}
    \includegraphics[width=0.98\textwidth]{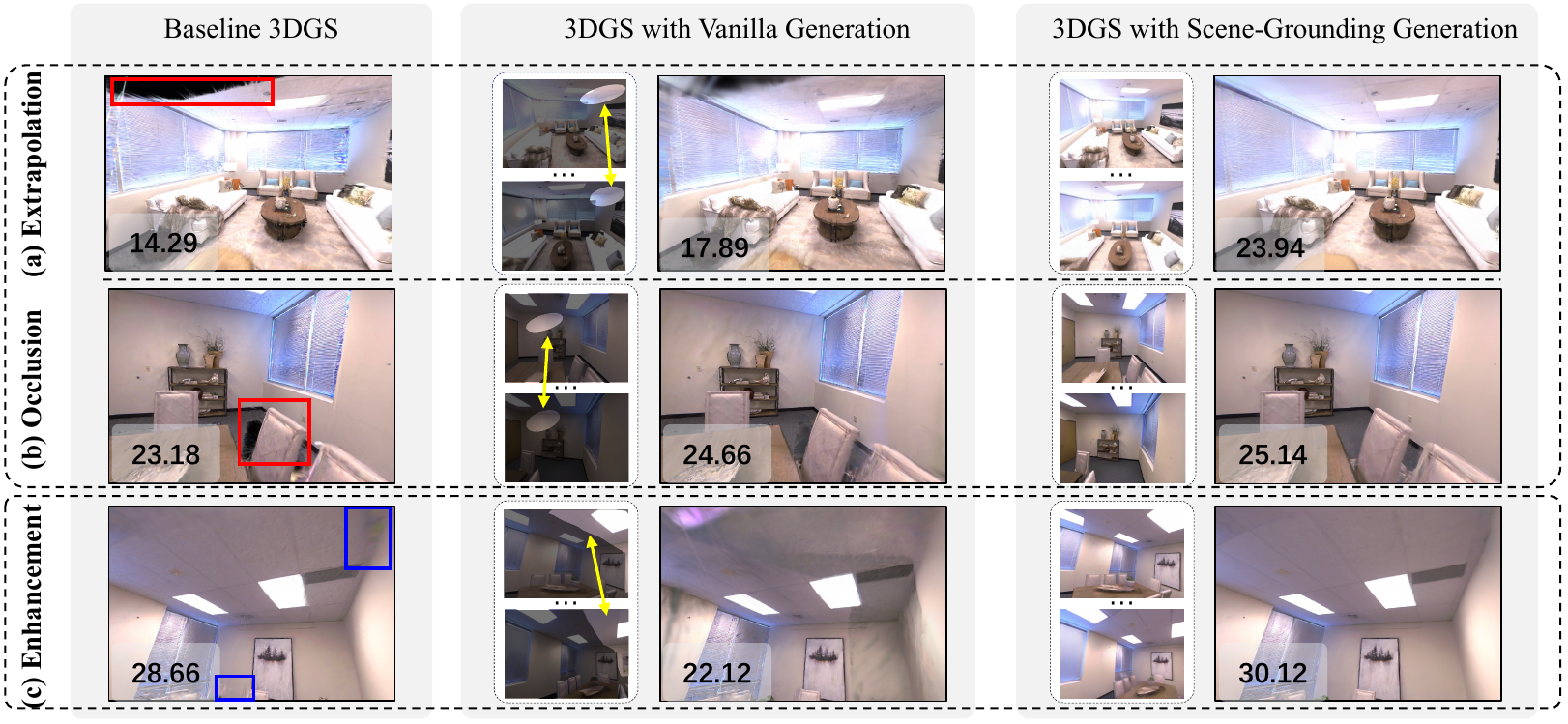}
    \vspace{-8pt}
    \captionof{figure}{We tackle the critical issues of \textbf{(a) extrapolation and (b) occlusion} in sparse-input 3DGS by leveraging a video diffusion model. Vanilla generation often suffers from inconsistencies within the generated sequences (as highlighted by the yellow arrows), leading to black shadows in the rendered images. In contrast, our scene-grounding generation produces consistent sequences, effectively addressing these issues and enhancing overall quality (c), as indicated by the blue boxes. The numbers refer to PSNR values. Zoom in for better~visualization. 
    }\vspace{-2pt}\label{fig:teaser}
\end{center}%
}]
\begin{abstract}
Despite recent successes in novel view synthesis using 3D Gaussian Splatting (3DGS), modeling scenes with sparse inputs remains a challenge. In this work, we address two critical yet overlooked issues in real-world sparse-input modeling: extrapolation and occlusion. To tackle these issues, we propose to use a reconstruction by generation pipeline that leverages learned priors from video diffusion models to provide plausible interpretations for regions outside the field of view or occluded. However, the generated sequences exhibit inconsistencies that do not fully benefit subsequent 3DGS modeling. To address the challenge of inconsistencies, we introduce a novel scene-grounding guidance based on rendered sequences from an optimized 3DGS, which tames the diffusion model to generate consistent sequences. This guidance is training-free and does not require any fine-tuning of the diffusion model. To facilitate holistic scene modeling, we also propose a trajectory initialization method. It effectively identifies regions that are outside the field of view and occluded. We further design a scheme tailored for 3DGS optimization with generated sequences. Experiments demonstrate that our method significantly improves upon the baseline and achieves state-of-the-art performance on challenging benchmarks. The project page is available at \href{https://zhongyingji.github.io/guidevd-3dgs/}{https://zhongyingji.github.io/guidevd-3dgs}. 
% In this work, we address two critical issues that are overlooked by current methods in real-world sparse-input modeling, i.e., extrapolation and occlusion.
% In this work, we focus on two issues that are overlooked by current works but critical for real-world sparse-input modeling, i.e., extrapolation and occlusion. 
% To address these issues, we intuitively propose a reconstruction-by-generation pipeline, in which we exploit learned priors from video diffusion models, providing plausible interpretations for regions that are outside field-of-view or occluded. 
% However, we observe that the generated sequences exhibit a certain level of inconsistency, which does not comprehensively benefit the subsequent 3DGS modeling. 
% To tackle the challenge of inconsistency, we propose a novel scene-grounding guidance, which is based on rendered sequences from an optimized 3DGS, to tame the diffusion model to directly generate consistent video sequences. 
% The designed guidance is training-free without including any finetuning of the diffusion model. 
% To enable a holistic modeling of the scene, we further propose a trajectory initialization method. 
% It effectively identifies regions that are outside field-of-view or occluded. Extensive experiments demonstrate that our method yields significant improvement over the baseline, and achieves state-of-the-art performance on challenging benchmarks. 

\end{abstract}
\vspace{-10pt}
\section{Introduction}
\label{sec:intro}
% nvs, nerf, 3dgs
Recent advances in 3D scene representation such as Neural Radiance Fields (NeRF)~\cite{mildenhall2020nerf,barron2021mip,verbin2022ref,barron2022mip,muller2022instant,barron2023zip} have greatly boosted the performance of Novel View Synthesis (NVS). NeRF represents the scene with a Multi-Layer Perceptron (MLP) and renders high-fidelity images with volumetric rendering. 
More recently, 3D Gaussian Splatting~(3DGS)~\cite{kerbl20233d,yu2024mip,lu2024scaffold,mallick2024taming} emerges as a powerful explicit representation that models the scene with a set of 3D gaussian primitives and renders images via differentiable splatting. 
3DGS achieves comparable performance to NeRF while requiring significantly less training time and offering higher inference speeds. 
% In the following, we collectively refer to NeRF and 3DGS as radiance fields.
\par Despite recent advances in scene representations based on 3DGS, modeling scenes with sparse inputs remains a significant challenge. The sparse supervision often leads radiance fields to learn degenerate representations due to shape-radiance ambiguity~\cite{zhang2020nerf++}, regardless of whether the representation is NeRF or 3DGS.
% Despite the progress of robust explicit scene representations, modeling a scene with sparse inputs remains a severe challenge, in which the sparse-view supervision makes the radiance fields learn a degenerate representation that is derived from shape-radiance ambiguity~\cite{zhang2020nerf++}, no matter if the underlying representation is NeRF or 3DGS. 
While there have been promising improvements~\cite{niemeyer2022regnerf,wang2023sparsenerf,zhong2024cvt,li2024dngaussian}, the commonly used face-forwarding~\cite{mildenhall2019local,jensen2014large} and object-oriented `outside-in' viewing~\cite{mildenhall2020nerf} settings oversimplify real-world sparse-input modeling, causing many methods to overlook two critical issues: 
\textbf{(i)} \textbf{extrapolation} - while the sparse inputs typically cover the scene as much as possible, there may still exist regions that are outside the field of view, as shown in Fig.~\ref{fig:teaser} (a); \textbf{(ii)} \textbf{occlusion} - occlusion frequently occurs for novel views that deviate even slightly from the training input views, 
as illustrated in Fig.~\ref{fig:teaser} (b). 
When rendering with an optimized 3DGS, these issues can cause severe artifacts, such as black holes, significantly degrading image quality.
% For rendering with an optimized 3DGS, these two issues will cause severe artifacts (e.g., black holes) in the respective regions and greatly degrade the quality of rendered images. 
Therefore, handling these two issues, i.e., extrapolation and occlusion, is critical for real-world sparse-input modeling. 
% These issues are even more severe in indoor scenes~\cite{}. 
% Note that this is not a generation problem, 

To address the above-discussed issues, we propose a novel reconstruction by generation pipeline based on 3DGS. Intuitively, we use video diffusion models~\cite{chen2023videocrafter1,ho2022imagen,xing2025dynamicrafter,yu2024viewcrafter} to generate multi-view sequences, which provide plausible interpretations of the scene based on priors learned from large-scale datasets. 
These sequences significantly enlarge the viewing instances, offering a high potential to address the extrapolation and occlusion issues. With the sparse inputs and the generated sequences, we can optimize a 3DGS to model the scene. However, as shown in Fig.~\ref{fig:teaser}, this vanilla pipeline brings little improvement or may even degrade the performance. The main reason can be attributed to the multi-view inconsistency within the generated sequences. 
The inconsistency manifests in two aspects: (\textbf{i}) the appearance inconsistency between frames within a sequence; (\textbf{ii}) the generated sequence may contain hallucinated elements not present in the scene.

\par 
% In this paper, we further explore fully leveraging the learned priors from video diffusion models for sparse-input 3DGS by addressing the challenges of inconsistency within the generated sequences.
% To fully leverage the learned priors from video diffusion models for sparse-input 3DGS, we further address the challenge of inconsistencies within the generated sequences.
To fully leverage the learned prior from video diffusion models for sparse-input 3DGS, we further explore addressing the challenges of inconsistencies within the generated sequences. 
Unlike existing methods that resolve appearance inconsistencies by assigning per-frame learnable appearance embeddings~\cite{martin2021nerf,lin2024vastgaussian}, we focus on taming video diffusion models to \emph{directly generate sequences with consistency}.
% Different from the existing methods of solving the appearance inconsistency by assigning the generated sequences with per-frame learnable appearance embeddings~\cite{martin2021nerf,lin2024vastgaussian}, we target at taming the video diffusion models to \emph{directly generate sequences with consistency.} 
Inspired by training-free guidance methods for diffusion models~\cite{bansal2023universal,yu2023freedom,song2023loss,ye2024tfg} that enable controllable generation through external guidance, we introduce a novel strategy called \emph{scene-grounding guidance} to ensure consistent generation without requiring further fine-tuning of the diffusion models.
% Inspired by training-free guidance methods for diffusion models~\cite{bansal2023universal,yu2023freedom,song2023loss,ye2024tfg} that perform controllable generation with external guidance, in this paper, {we tame the video diffusion models with a novel strategy of \emph{scene-grounding guidance} to control the generation to be consistent, without any further finetuning of the diffusion models.} 
Specifically, the scene-grounding guidance is based on a rendered sequence from an optimized 3DGS. During each step of the denoising process, the noisy sequence receives gradients from the supervision of the rendered sequence.
% The gradient is involved into the denoising sampler, and each denoising step thus considers the guidance from the rendered sequence. 
Although the rendered sequence does not provide perfect guidance, our key insight in employing it to address the inconsistency is twofold: \textbf{(i)} adjacent frames within the rendered sequence are highly consistent due to limited camera movement between them;
% Although the rendered sequence does not provide perfect guidance, our key insight of employing it to address the inconsistency is: (\textbf{i}) the adjacent frames within the rendered sequence are highly consistent, due to the limited camera movement between them; 
(\textbf{ii}) the rendered sequence is scene-grounding, which can guide the diffusion model to avoid generating elements that do not exist in the scene. In addition to addressing the issues of extrapolation and occlusion, our method also enhances the overall quality of the rendered images, as demonstrated in Fig.~\ref{fig:teaser} (c). 
To effectively identify regions that are outside the field of view or occluded, we propose a trajectory initialization strategy to determine the camera trajectory during sequence generation, which is also based on an optimized 3DGS. 
% For each sparse input view, we first define a set of candidate poses around it. Using the optimized 3DGS, we render images for these candidate poses. For poses where the rendered images display significant holes, we interpolate a trajectory between the input view and these poses. 
With the proposed method, we can perform a holistic modeling of the scene. 
% the generated sequences can cover the scene to the largest extent, boosting the overall quality of scene modeling. 
% Our method 
% without any finetuning, thus is universal as it is theoretically applicable to different video diffusion models. 
Additionally, we introduce a scheme for optimizing 3DGS with generated sequences, focusing on loss and sampling designs, which further enhance overall performance.
Following~\cite{zhong2025empowering}, we evaluate our method on two challenging indoor datasets: Replica~\cite{straub2019replica} and ScanNet++~\cite{yeshwanth2023scannet++}, where the issues of extrapolation and occlusion are pronounced. The experiments demonstrate that our method achieves notable improvements and establishes state-of-the-art performance. Our contributions are summarized as: 
\begin{itemize}
    \item This paper is the first to explicitly address the challenges of extrapolation and occlusion in 3DGS modeling from sparse inputs. 
    \item We propose a~novel~reconstruction by generation pipeline with a designed scene-grounding guidance, which tames the video diffusion models to generate consistent and plausible sequences to effectively tackle the issues of extrapolation and occlusion.  
    \item We present a trajectory initialization strategy that effectively identifies regions that are outside the field of view and occluded, facilitating holistic scene modeling. 
    We also introduce a scheme for optimizing 3DGS with generated sequences, further improving the performance. 
    \item Our method demonstrates significant improvements over the baseline, achieving over 3.5 dB and 2.5 dB PSNR enhancements on the Replica~\cite{straub2019replica} and ScanNet++~\cite{yeshwanth2023scannet++} datasets, respectively, thereby establishing state-of-the-art performance. 
\end{itemize}

% however, sparse view xxx. although xxx, current method cannot tackle indoor scene, difficulties

% In this paper, we use video diffusion. Why? great understanding of the physical world. It can xxx 3 improvement. However, xxx, as shown in Fig. xxx

% Our method 
% different with current method that tries to solve the inconsistencies like vastgaussian xxx, nerfw, xxx
% why 
% implementation

\section{Related Works}
\label{sec:related}
% \noindent \textbf{Radiance Fields for Novel View Synthesis. }
% Neural Radiance Fields (NeRF)~\cite{mildenhall2020nerf,barron2021mip,verbin2022ref,liu2020neural,barron2022mip,muller2022instant,barron2023zip} have achieved impressive novel view synthesis results in recent years. NeRF trains a Multi-Layer Perceptron (MLP) to model the appearance and the geometry of the scene, and images are rendered through volumetric rendering. 
% Recently, 3D Gaussian Splatting (3DGS)~\cite{kerbl20233d,yu2024mip,lu2024scaffold,niemeyer2024radsplat,mallick2024taming} further pushes the boundary of scene representation, achieving not only high-fidelity synthesis results, but also significantly decreased inference time. 3DGS represents the scene with a set of discretized 3D gaussian primitives, which renders an image through a differentiable splatting.  

\begin{figure*}[t]
  \centering
  \includegraphics[width=0.95\linewidth]{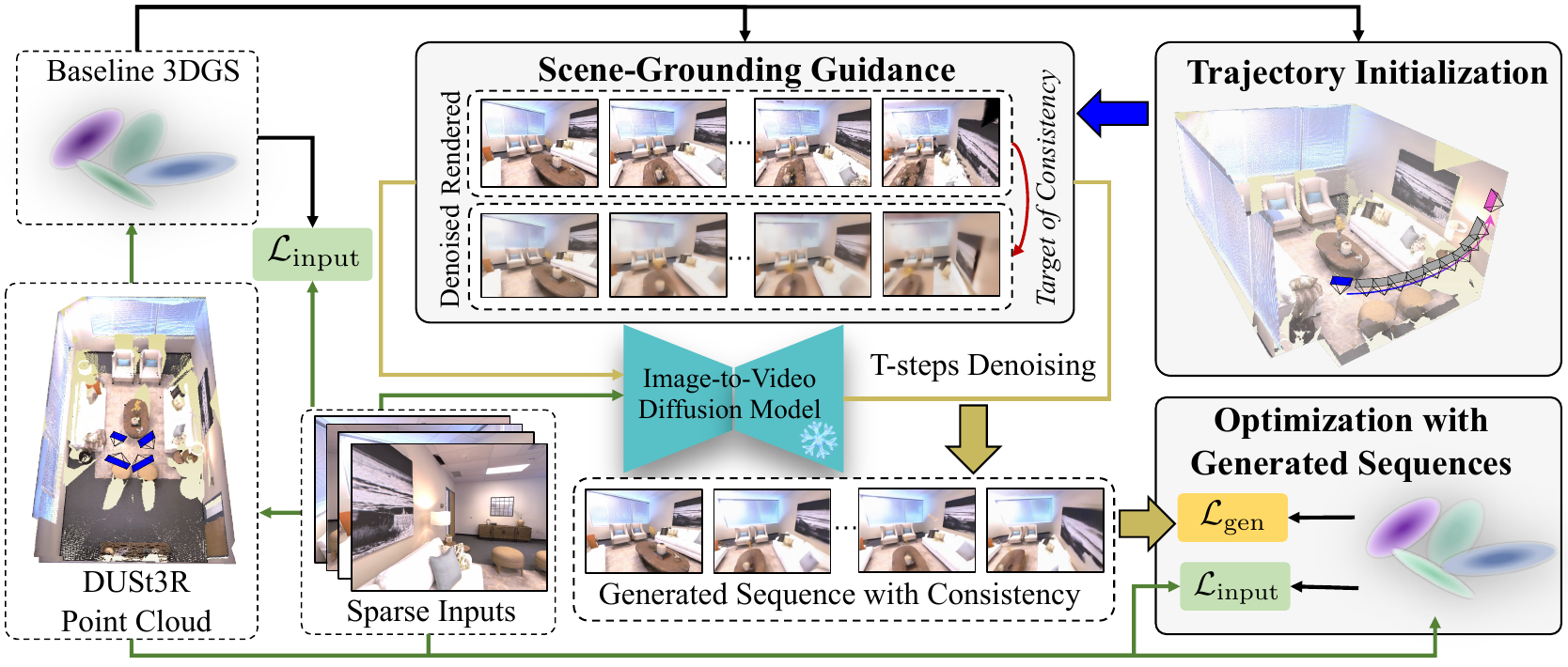}
  \vspace{-8pt}
  \caption{\textbf{Framework overview of our proposed method.} It consists of three parts: scene-grounding guidance, trajectory initialization, and optimization scheme with generated sequences. 
  Initially, a baseline 3DGS is trained using sparse inputs and initialized with the point cloud from DUSt3R~\cite{wang2024dust3r}. 
  Yellow regions denote uncovered areas, e.g., those outside the field of view or occluded.
  The trajectory initialization determines the paths for sequence generation based on renderings from the baseline 3DGS, facilitating holistic scene modeling. The video diffusion model receives an input image along with the trajectory for sequence generation, incorporating scene-grounding guidance during the denoising process to ensure consistent output. The guidance is based on the rendered sequences. 
  Finally, the generated sequences are utilized to optimize the final 3DGS through a tailored optimization scheme. 
  }\vspace{-0pt}
  \label{fig:framework}
\end{figure*}

\noindent \textbf{Radiance Fields from Sparse Inputs. }Although improvements have been made in scene representation~\cite{mildenhall2020nerf,barron2021mip,barron2022mip,muller2022instant,barron2023zip,kerbl20233d,yu2024mip,lu2024scaffold}, learning a robust radiance field typically requires dense inputs as supervision due to the radiance-shape ambiguity~\cite{zhang2020nerf++}. 
Current works can be roughly categorized into two lines of research. The first line of research focuses on pre-training generalizable radiance fields using multi-view datasets. Generalizable NeRF~\cite{yu2021pixelnerf,wang2021ibrnet,liu2022neural,chen2021mvsnerf} typically projects the ray points to reference images and integrates 2D features as the auxiliary input of MLP. In contrast, generalizable 3DGS~\cite{liu2025mvsgaussian,charatan2024pixelsplat,chen2024mvsplat,zhang2025gs} commonly perform dense depth prediction associated with gaussian properties. 
The second line of research leverages regularization techniques for optimization, which are applicable to both NeRF and 3DGS, including hand-crafted constraints~\cite{kim2022infonerf,niemeyer2022regnerf,zhong2024cvt}, and those derived from pre-trained models~\cite{radford2021learning,wang2023sparsenerf,wynn2023diffusionerf,li2024dngaussian,zhu2025fsgs,fan2024instantsplat,paliwal2025coherentgs}.
In this paper, we explore the priors from video diffusion models~\cite{chen2023videocrafter1,ho2022imagen,xing2025dynamicrafter,yu2024viewcrafter} for sparse-input 3DGS modeling.
% In this paper, we explore to use the strong prior from video diffusion models~\cite{chen2023videocrafter1,ho2022imagen,xing2025dynamicrafter,yu2024viewcrafter} for sparse-input 3DGS modeling.
Unlike previous works, our method trains sparse-input 3DGS using augmented sequences from video diffusion models, which provide interpolation or extrapolation around the input views. We design a guidance strategy that tames the video diffusion model to generate more plausible and scene-grounded sequences, greatly enhancing the performance. 
% The most relevant method with ours is ReconX~\cite{liu2024reconx}, which trains a 3D-aware video diffusion model and uses the generated sequences to train a 3DGS with uncertainty prediction. 
% However, we observe that 

\noindent \textbf{Diffusion Prior for Radiance Fields. }Diffusion models~\cite{song2019generative,ho2020denoising,rombach2022high} have shown remarkable generation capabilities. The strong prior knowledge embedded in diffusion models can facilitate the training of radiance fields.
Specifically, several works leverage Score Distillation Sampling (SDS)~\cite{poole2022dreamfusion,lin2023magic3d,wang2024prolificdreamer,liang2024luciddreamer,chen2024scenetex} using a frozen diffusion model to train a 3D consistent representation based on text prompts in a zero-shot manner.
% Concretely, certain amount of works leverage 
% Score Distillation Sampling (SDS)~\cite{poole2022dreamfusion,lin2023magic3d,wang2024prolificdreamer,liang2024luciddreamer} with a frozen diffusion model to train a radiance field with text prompt in a zero-shot manner.
Other studies focus on training view-consistent~\cite{shi2023mvdream,tang2024mvdiffusion++,wu2024reconfusion} or quality-enhanced~\cite{liu20243dgs} diffusion models, where the generated results can be directly applied to train a radiance field.
% Some other works focus on training view-consistent diffusion models~\cite{shi2023mvdream,tang2024mvdiffusion++,wu2024reconfusion}, whose generated results can be directly applied to train a radiance field. 
Our method also uses the generated results to train a radiance field. 
However, our primary contribution lies in a novel guidance strategy that controls the generation to be consistent, which is crucial for sparse-input modeling augmented with generation.
% However, our main contribution lies in a guidance strategy that controls the generation process, which is of great significance for sparse-input modeling augmented with generated results. 

\noindent \textbf{Controllable Generation for Diffusion Models. }
% Researchers are studying effective ways to control the generation process of powerful diffusion models. 
Current works of controllable generation can be categorized into training-required and training-free methods. Training-required methods fine-tune the diffusion models with additional conditions~\cite{zhang2023adding,mou2024t2i}, 
% e.g., ControlNet~\cite{zhang2023adding} and T2I-Adapter~\cite{mou2024t2i}, 
or train an additional noise-conditioned external guidance function, e.g., classifier guidance~\cite{dhariwal2021diffusion} for denosing sampler. 
Training-free methods freeze the foundation diffusion model, and modify the denoising process with the control signal from the external guidance functions~\cite{bansal2023universal,yu2023freedom,song2023loss,ye2024tfg}. These methods do not require training additional guidance functions or fine-tuning diffusion models; instead, they enable controllable generation in a plug-and-play manner. 
Our work is inspired by training-free methods. However, we concentrate on multi-view modeling, which necessitates high consistency control over the generated results. Besides, unlike these methods that typically rely on pre-trained models for guidance, we utilize rendered results to provide guidance.
\section{The Proposed Method}
In this paper, we utilize video diffusion models to tackle two critical issues in real-world sparse-input modeling: extrapolation and occlusion, as illustrated in Fig.~\ref{fig:teaser}.
The overview of our method is illustrated in Fig.~\ref{fig:framework}, which consists of three proposed components: a scene-grounding guidance (Sec.~\ref{sec:guidance}), a trajectory initialization strategy (Sec.~\ref{sec:traj}), and a scheme for 3DGS optimization with generated sequences (Sec.~\ref{sec:train_gs}). 
We will detail these components following a preliminary review of 3DGS and diffusion models.

\subsection{Preliminary}
\textbf{3D Gaussian Splatting} (3DGS)~\cite{kerbl20233d} represents a scene with a set of anisotropic 3D Gaussian primitives. Each Gaussian primitive is parametrized by a set of attributes: a center $\mu$, a scaling factor $s$, a quaternion $q$, an opacity value $\alpha$, and a feature vector $f$.~The basis function of each Gaussian primitive is formulated as $\mathcal{G}(x)=\rm{exp}({-\frac{1}{2}(x-\mu)^T \Sigma^{-1}(x-\mu)})$, where $\Sigma$ is the covariance matrix derived from $s$ and $q$. 3DGS renders the scene through a differentiable splatting, which firstly transforms the 3D Gaussian $\mathcal{G}(x)$ into 2D Gaussian $\mathcal{G}^{'}(x)$ on the image plane via projection~\cite{zwicker2002ewa}, and applies a tile-based rasterizer for rendering, which sorts the 2D Gaussians by depth and employs the $\alpha$-blending as follows: 
\begin{equation}\label{eq:splat}
\setlength{\abovedisplayskip}{0.05cm}
\setlength{\belowdisplayskip}{0.02cm}
C(x_p) = \sum_{i \in K}c_i \sigma_i \prod_{j=1}^{i-1}(1-\sigma_j), \quad \sigma_i = \alpha_i \mathcal{G}^{'}_{i}(x_p), 
\end{equation}
where $x_p$ is the pixel position, $K$ refers to the number of 2D Gaussians associated with the pixel, and $c$ represents the decoded color of feature $f$. 

\noindent \textbf{Diffusion Models}~\cite{song2019generative,ho2020denoising} are a family of generative models that progressively perturb data with intensifying Gaussian noises (i.e., forward noising), and then learn to reverse this process for sample generation (i.e., reverse denoising). The key of the diffusion model is a U-Net $\epsilon_{\theta}$ which is trained to predict the noise that is injected in the current sample $
\mathbf{x}_t$. The sampling is conducted by iterative denoising for $T$ steps~\cite{yu2023freedom} as follows: 
% \begin{equation}\label{eq:ddpm}
% \textbf{x}_{t-1}=\frac{1}{\sqrt{\alpha_t}}(\textbf{x}_t-\frac{1-\alpha_t}{\sqrt{1-\bar{\alpha}_t}}\epsilon_{\theta}(\textbf{x}_t, t))+\sigma_t\textbf{z}, 
% \end{equation}
\begin{equation}\label{eq:ddpm}
    \mathbf{x}_{t-1}=(1+\beta_{t}/2)\mathbf{x}_t+\beta_t \nabla_{\mathbf{x}_t}\log p({\mathbf{x}_t}) + \sqrt{\beta_t}\mathbf{z}
\end{equation}
where $\nabla_{\mathbf{x}_t}\log p({\mathbf{x}_t})$ is the estimated score function which can be derived from $\epsilon_{\theta}(\mathbf{x}_t, t)$; $\beta_t$ is pre-defined parameters; $\mathbf{z}\sim\mathcal{N}(0, \mathbf{I})$. 
In this work, we leverage a camera-controlled image-to-video diffusion model~\cite{yu2024viewcrafter}, whose condition includes an image for the first frame, and the camera trajectory for the path of the generated sequence. 
% In the following, we omit the condition in $\epsilon_{\theta}$ to avoid being cluttered. 
The model is operated in a latent space of dimension $d$, supporting the sequence length of $L$, thus $\mathbf{x}_t \in \mathbb{R}^{L\times h\times w\times d}$.

% \section{The Proposed Method}
% In this paper, we use video diffusion models to address two issues that are critical for real-world sparse-input modeling, i.e., extrapolation and occlusion, as shown in Fig.~\ref{fig:teaser}. 
% Our method is illustrated in Fig~\ref{fig:framework}, which comprises three parts, i.e., a scene-grounding guidance (Sec.~\ref{sec:guidance}), a trajectory initialization method (Sec.~\ref{sec:traj}), and strategies for 3DGS optimization with generated sequences (Sec.~\ref{sec:train_gs}). We will introduce their details in the following.  

\subsection{Generation via Scene-Grounding Guidance}\label{sec:guidance}
Applying the generated sequences from the video diffusion model can provide plausible interpretations of regions not covered by the sparse inputs. However, as illustrated in Fig.~\ref{fig:teaser}, the inconsistency within the generated sequences manifests as: (\textbf{i}) appearance inconsistencies across frames and (\textbf{ii}) the occurrence of non-existent elements, which can negatively impact the 3DGS modeling. 
In this section, we propose an innovative scene-grounding guidance method that directs the video diffusion model to generate consistent sequences, significantly enhancing the performance of sparse-input 3DGS.

Inspired by previous training-free guidance methods~\cite{bansal2023universal,yu2023freedom} that achieve their objectives by modifying the sampler in Eq.~\eqref{eq:ddpm}, we adopt a similar approach to attain the goal of consistency. Specifically, we firstly replace $\nabla_{\mathbf{x}_t}\log p({\mathbf{x}_t})$ with a conditional score function $\nabla_{\mathbf{x}_t}\log p({\mathbf{x}_t}\vert \mathcal{Q})$, 
where $\mathcal{Q}$ refers to \emph{the target of consistency.} The conditional score function can be expanded by the Bayesian rule as: 
\begin{equation}\label{eq:bayesian}
\nabla_{\mathbf{x}_t}\log p({\mathbf{x}_t}\vert \mathcal{Q})=\nabla_{\mathbf{x}_t}\log p({\mathbf{x}_t})+\nabla_{\mathbf{x}_t}\log p({\mathcal{Q}\vert\mathbf{x}_t}), 
\end{equation}
where $\nabla_{\mathbf{x}_t}\log p({\mathcal{Q}\vert\mathbf{x}_t})$ can be considered as a guidance term that injects the consistency constraint into Eq.~\eqref{eq:ddpm}. We further formulate $p({\mathcal{Q}\vert\mathbf{x}_t})$ as: $p({\mathcal{Q}\vert\mathbf{x}_t})={\rm{exp}}(-\lambda \mathcal{L}(\mathcal{Q}, \mathbf{x}_t))/Z$, where $\mathcal{L}(\mathcal{Q}, \mathbf{x}_t)$ indicates how well the current sample $\mathbf{x}_t$ is aligned with the target, and $Z$ is a normalization term. The guidance term can thus be implemented using the gradient of the following loss function: 
\begin{equation}\label{eq:guide_term}
\nabla_{\mathbf{x}_t}\log p({\mathcal{Q}\vert\mathbf{x}_t}) \propto -\nabla_{\mathbf{x}_t}\mathcal{L}(\mathcal{Q}, \mathbf{x}_t),  
\end{equation}
which is appended to Eq.~\eqref{eq:ddpm} to achieve the target of consistency during the denoising sampling. 

\begin{figure}[t]
\vspace{-1em}
\begin{algorithm}[H]
\caption{Generation with Scene-Grounding Guidance}
\label{alg:guide}
\begin{algorithmic}[1]
\STATE \textbf{Function} GENERATOR($\mathcal R$, $I$, $\{\phi_{j}\}_{j=1}^L$)
\STATE \textbf{Input: }Optimized 3DGS model $\mathcal{R}$, input image $I$, camera trajectory of a sequence $\{\phi_{j}\}_{j=1}^L$. 
\STATE \textbf{Given: }Latent image-to-video diffusion model $\epsilon_{\theta}$, VAE decoder $\mathcal{D}$, pre-defined $\beta_t, \bar{\alpha}_t$ and guidance scale $\gamma_t$.
\STATE \textbf{Abbreviate} $\epsilon_{\theta}(\mathbf{x}_t, t, I, \{\phi_{j}\}_{j=1}^L)$ \textbf{as} $\epsilon_{\theta}(\mathbf{x}_t, t)$
\STATE $\mathbf{S}, \mathbf{M}={\rm{rasterize}}(\{\phi_{j}\}_{j=1}^L, \mathcal{R})$ \hfill $\triangleright$ Eq.~\eqref{eq:splat}\&~\eqref{eq:mask}
\STATE $\mathbf{x}_T \sim \mathcal{N}(0, \mathbf{I})$
\FOR{$t = T, \ldots, 1$}
    \STATE $\mathbf{z} \sim \mathcal{N}(0, \mathbf{I})$ if $t > 1$, else $\mathbf{z} = \mathbf{0}$
    \STATE $\hat{\mathbf{x}}_{t-1} = (1 + \frac{1}{2}\beta_t)\mathbf{x}_t - \frac{\beta_t}{\sqrt{1-\bar{\alpha}_t}} \epsilon_{\theta}(\mathbf{x}_t, t) + \sqrt{\beta_t} \mathbf{z}$
    \STATE $\mathbf{x}_{0|t} = \frac{1}{\sqrt{\bar{\alpha}_t}}(\mathbf{x}_{t} - \sqrt{1 - \bar{\alpha}_t} \epsilon_{\theta}(\mathbf{x}_t, t))$
    \STATE $\mathbf{X}_{0|t} = \mathcal{D}(\mathbf{x}_{0|t})$
    \STATE $\mathbf{g}_t = \nabla_{\mathbf{x}_t} \mathcal{L} (\mathbf{S}, \mathbf{M}, \mathbf{X}_{0|t})$ \hfill $\triangleright$ Eq.~\eqref{eq:guide_loss}
    \STATE $\mathbf{x}_{t-1} = \hat{\mathbf{x}}_{t-1} - \gamma_t \mathbf{g}_t$ \hfill $\triangleright$ Eq.~\eqref{eq:ddpm}\&~\eqref{eq:guide_term}
\ENDFOR
\RETURN $\mathcal{D}(\mathbf{x}_0)$
\end{algorithmic}
\end{algorithm}
\vspace{-8mm}
\end{figure}

The remaining problem lies in how to define the consistency target $\mathcal{Q}$. 
Unlike previous works~\cite{bansal2023universal,yu2023freedom,song2023loss} that define the target based on external pre-trained models, we establish the target using a rendered sequence from an optimized 3DGS model $\mathcal R$.
% Different with previous works~\cite{bansal2023universal,yu2023freedom,song2023loss} that defines the target based on external pre-trained models, we define the target based on a rendered sequence from an optimized 3DGS model $\mathcal R$. 
Though the rendered sequence is not perfect, our key insights are as follows: \textbf{(i)} the rendered images of adjacent frames are highly consistent, as the camera movement between them is typically minor; 
\textbf{(ii)} the rendered frames provide scene grounding, clearly indicating which elements exist in the scene. Therefore, the rendered sequence can serve as an effective guidance for the generated sequence to achieve the target of consistency. 

Given a camera trajectory $\{\phi_{j}\}_{j=1}^L$ for the sequence generation, we first utilize the optimized 3DGS to render a sequence $\{S_j\}_{j=1}^L$, 
%(stacked into $\mathbf{S}\in \mathbb{R}^{L\times H\times W\times 3}$)
along with a mask sequence $\{M_j\}_{j=1}^L$ that indicates the regions not covered by the sparse inputs. To get the mask, we first render a transmittance map, which is obtained by $\alpha$-blending (as Eq.~\eqref{eq:splat}) on the opacity. For each pixel $x_p$, the $\alpha$-blending is formulated as: 
\begin{equation}\label{eq:mask}
\setlength{\abovedisplayskip}{0.09cm}
\setlength{\belowdisplayskip}{0.05cm}
O(x_p) = \sum_{i \in K}\sigma_i \prod_{j=1}^{i-1}(1-\sigma_j),  
\end{equation}
where $\sigma$ and $O$ refer to the opacity of the gaussian and the transmittance map, respectively. The mask is then acquired by thresholding the transmittance map with a value $\eta_{\rm{mask}}$: $M=\left(O<\eta_{\rm{mask}}\right)$. 
For convenience, we stack $\{S_j\}_{j=1}^L$ and $\{M_j\}_{j=1}^L$ to $\mathbf S$ and $\mathbf M$, which are of shape $L\times H\times W\times 3$ and $L\times H\times W\times 1$, respectively. 
Since the target of consistency is based on the rendered sequence in clean data space, to receive the guidance, we transform the noisy latent $\mathbf{x}_t$ into a latent $\mathbf{x}_{0\vert t}$ in the clean data space, based on prediction from the model $\epsilon_{\theta}$: 
$\mathbf{x}_{0|t} = (\mathbf{x}_{t} - \sqrt{1 - \bar{\alpha}_t} \epsilon_{\theta}(\mathbf{x}_t, t))/\sqrt{\bar{\alpha}_t}.$
 With the consistency target $\mathcal{Q}$ that is based on the rendered sequence $\mathbf{S}$, we formulate the function $\mathcal{L}$ in the guidance term (Eq.~\eqref{eq:guide_term}) as: 
\begin{equation}\label{eq:guide_loss}
\begin{aligned}
\mathcal{L}(\mathbf{S}, \mathbf{M}, \mathbf{X}_{0|t})&=\Vert \mathbf{M} \odot (\mathbf{S}-\mathbf{X}_{0\vert t}) \Vert_{1} \\
& + \lambda_{\rm{perc}}\mathcal{L}_{\rm{perc}}(\mathbf{M} \odot \mathbf{S}, \mathbf{M} \odot \mathbf{X}_{0\vert t}), 
\end{aligned}
\end{equation}
% + \lambda_{1}\mathcal{L}_{\rm{perc}}(\mathbf{S}, \mathbf{X}_{0\vert t})
where $\mathbf{X}_{0\vert t}$ is decoded from the latent $\mathbf{x}_{0\vert t}$ by a VAE decoder, $\odot$ is the Hadamard product, and $\mathcal{L}_{\rm{perc}}$ is a perceptual loss~\cite{johnson2016perceptual} with its corresponding weight as $\lambda_{\rm{perc}}$.  

% some conclusion
With the guidance from Eq.~\eqref{eq:guide_loss}, the denoising process balances the consistency constraint and the prior from the diffusion model, integrating them into plausible generation results. 
This guidance does not involve any fine-tuning of the diffusion model, thereby preserving its generative capabilities. 
% Since the guidance is based on the scene-grounding renderings from 3DGS, we refer to it as scene-grounding guidance. 
The detailed pipeline is outlined in Alg.~\ref{alg:guide}.

\subsection{Trajectory Initialization Strategy}\label{sec:traj}
\begin{figure}[!t]
  \centering
  \includegraphics[width=0.99\linewidth]{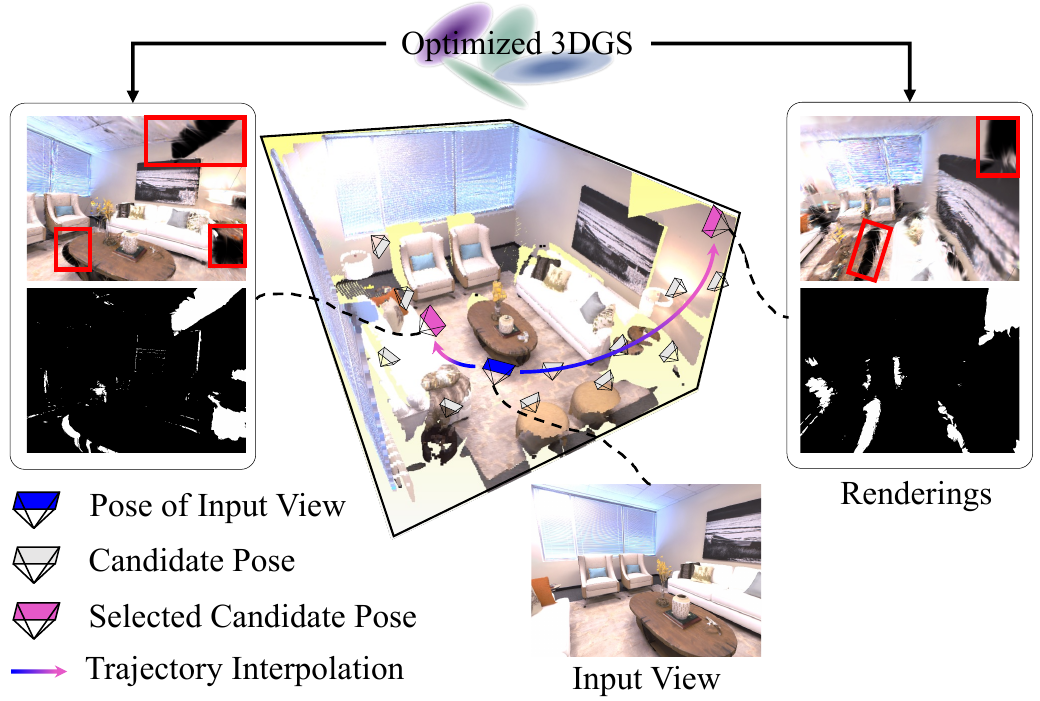}
  \vspace{-0mm}
  \caption{Illustration of the proposed trajectory initialization strategy. The yellow parts represent unobserved regions. For each input view, we sample a set of candidate poses around it, and render at these poses using an optimized 3DGS. We select candidate poses whose renderings exhibit significant holes (highlighted by red boxes), and interpolate trajectories between these candidate poses and the input view's pose.}
  \vspace{-4mm}
  \label{fig:traj}
\end{figure}

% \subsection{Trajectory Initialization}\label{sec:traj}
To enable holistic modeling of the scene, the camera trajectories for the video diffusion model should cover regions that are outside the field of view or occluded as much as possible.
The generated sequences can thus provide plausible interpretations for these regions, which serve as the basis for optimizing the subsequent 3DGS model. 
Similar to the scene-grounding guidance discussed in Sec.~\ref{sec:guidance}, the proposed trajectory initialization method is also based on an optimized 3DGS model. 
For the $i$-th sparse input view with camera pose $\varphi_i$, we first sample a set of candidate poses around it, as depicted in Fig.~\ref{fig:traj}. Suppose there are a total of $m$ candidate poses, we use the optimized 3DGS model $\mathcal{R}$ to render images for these poses as $\{\hat{S}^{(i)}_c, \hat{M}^{(i)}_c\}_{c=1}^m = {\rm{rasterize}}(\{\hat{\phi}^{(i)}_c\}_{c=1}^m,   \mathcal{R})$.
For poses where the rendered images exhibit significant black holes, as indicated by the mask $\hat{M}$ calculated from Eq.~\eqref{eq:mask}, we interpolate a trajectory of length $L$ (matching the length of video diffusion model) between the input camera pose and these poses as follows: 
% \begin{equation}\label{eq:traj_interp}
% \setlength{\abovedisplayskip}{0.03cm}
% \setlength{\belowdisplayskip}{0.03cm}
$\{\phi_j^{(i, c)}\}_{j=1}^L = {\rm{interp}}(\varphi_i, \hat{\phi}_c^{(i)})$, 
% \end{equation}
where $\hat{\phi}_c^{(i)}$ refers to one selected candidate pose from the $i$-th input. 
In practice, we select the top-$k$ candidate poses based on the sizes of their corresponding masks. 
% Eq.~\eqref{method:traj_interp} represents a trajectory interpolated between pose of one input view and one selected candidate pose. 
Then, we build a trajectory pool by traversing all input views and their respective selected candidate poses as: 
\begin{equation}\label{eq:traj_pool}
\Phi = \{\{\phi_j^{(i, c)}\}_{j=1}^L\vert\ i, c\}, 
\end{equation}
where each element in the pool is sampled for the sequence generation.  

\begin{figure}[t]
\vspace{-1em}
\begin{algorithm}[H]
\caption{3DGS Optimization with Generation}
\label{alg:optim}
\begin{algorithmic}[1]
\STATE \textbf{Input: }Sparse inputs of N images $\{C_i^{\rm{gt}}, \varphi_i\}_{i=1}^N$. 
\STATE \textbf{Given: }Number of iterations $N_{\rm{iter}}$, generation interval $N_{\rm{gen}}$, ratio of samples from other sequences $\eta$. 
\STATE \textbf{Variable: }Global list of generated views $\mathbf{G}=[\,]$. 
\STATE Baseline 3DGS model optimization $\Rightarrow \mathcal R$
\STATE Trajectory initialization $\Rightarrow \Phi$ \hfill $\triangleright$ Eq.~\eqref{eq:traj_pool}
\FOR{$t = 0, \ldots, N_{\rm{iter}}-1$}
    \STATE \textbf{If} $t$ \% $N_{\rm{gen}}$ = 0 \textbf{then}
    \STATE \quad Sample an input view $I$
    \STATE \quad Sample a trajectory around $I$ from $\Phi \Rightarrow \{\phi_j\}_{j=1}^L$
    \STATE \quad $\mathbf{S}$ = ${\rm{GENERATOR}}(\mathcal R, I, \{\phi_j\}_{j=1}^L)$
    \STATE \quad Append $\mathbf{S}$ to $\mathbf{G}$
    \STATE \textbf{End If}
    \STATE Sample an input view to get $\mathcal{L}^{\rm{input}}$ \hfill $\triangleright$ Eq.~\eqref{eq:loss_input}
    % \STATE $r = \rm{rand}(\,)$
    \STATE \textbf{If} $\rm{rand}()\geq\eta$ \textbf{then} \STATE \quad Sample a generated view from $\mathbf{S}$
    \STATE \textbf{Else} Sample a generated view from $\mathbf{G}$
    \STATE \textbf{End If}
    \STATE Use the generated view to get $\mathcal{L}^{\rm{gen}}$ \hfill $\triangleright$ Eq.~\eqref{eq:loss_gen}
    \STATE ($\mathcal{L}^{\rm{input}}+\mathcal{L}^{\rm{gen}}$).backward(\,)
    \STATE \color{lightgray}\# Densification and opacity reset
\ENDFOR
\end{algorithmic}
\end{algorithm}
\vspace{-8mm}
\end{figure}

% \subsection{3DGS Optimization with Diffusion Generation}\label{sec:train_gs}
% \subsection{3DGS Optimization Scheme with Generation}\label{sec:train_gs}
\subsection{3DGS Optimization with Generation}\label{sec:train_gs}
Given sparse inputs of $N$ images along with their poses, i.e., $\{C_i^{\rm{gt}}, \varphi_i\}_{i=1}^N$, we aim at optimizing a 3DGS model with the auxiliary generated sequences. 
For simplicity, we refer to the input images paired with their poses as `input views', and we term the generated images with their associated poses as `generated views'. During each iteration, we sample an input view and a generated view for supervision. Specifically, for the input view, we employ the default reconstruction loss~\cite{kerbl20233d} written as: 
\begin{equation}\label{eq:loss_input}
    \mathcal{L}^{\rm{input}}=(1-\lambda)\mathcal{L}_1(C_i, C_i^{\rm{gt}}) + \lambda \mathcal{L}_{\rm{D-SSIM}}(C_i, C_i^{\rm{gt}}), 
\end{equation}
where $C_i$ refers to the rendered image and $\lambda$ is a weighting factor. 
For supervision of generated views, we find that the reconstruction loss does not effectively fill the hole regions, and increasing its weight leads to performance degradation due to flaws in the generated images. To address this issue, we propose using perceptual loss~\cite{johnson2016perceptual}. The perceptual loss is calculated over the entire image, allowing those hole regions to significantly influence the gradients, thereby effectively driving the model to fill those holes. 
Thus, the loss on the generated views is formulated as: 
\begin{equation}\label{eq:loss_gen}
\mathcal{L}^{\rm{gen}}=\lambda_{\rm{gen1}}\mathcal{L}_1(C_j, S_j) + \lambda_{\rm{gen2}} \mathcal{L}_{\rm{perc}}(C_j, S_j), 
\end{equation}
where $S_j$ refers to the generated image, $\lambda_{\rm{gen1}}$ and $\lambda_{\rm{gen2}}$ are two balancing factors, respectively.

\par We empirically find that conducting local sampling within a specific optimization interval, where a substantial portion of generated views is sampled from the same sequence of local regions, enhances visual quality.
However, sampling exclusively from a single sequence can lead to a forgetting issue, where optimized information about holes in other regions becomes diluted. 
Therefore, within each interval of local sampling, we also include generated views from other sequences with a ratio $\eta$. 
The optimization pipeline is presented in Alg.~\ref{alg:optim}.

\vspace{-6pt}
\section{Experiments}
\subsection{Experimental Setups}
\noindent \textbf{Datasets and Metrics. }
We target addressing the issues of extrapolation and occlusion for sparse-input 3DGS scene modeling, which are overlooked by current benchmarks. 
% which are commonly overlooked by current face-fowarding~\cite{mildenhall2019local,jensen2014large} and object-oriented ``outside-in''~\cite{mildenhall2020nerf} viewing benchmarks. 
To evaluate the effectiveness of our method, we conduct experiments on a benchmark~\cite{zhong2025empowering} created from two indoor datasets, i.e., the synthetic Replica~\cite{straub2019replica} and the realistic ScanNet++~\cite{yeshwanth2023scannet++}, which consists of 6 and 4 scenes, respectively. Although the selected six input views for each scene can cover most regions, there are still areas outside the field of view. Moreover, the `inside-out' viewing directions make occlusion common in this benchmark. 
For quantitative comparisons, we report PSNR, SSIM~\cite{wang2004image}, and LPIPS~\cite{zhang2018unreasonable} scores. 

\noindent \textbf{Baseline. }We train a baseline 3DGS model initialized with the point cloud from DUSt3R~\cite{wang2024dust3r}, 
% Our baseline 3DGS model is built upon the state-of-the-art method FSGS~\cite{zhu2025fsgs}, which is a strong baseline.
incorporating the gaussian unpooling in FSGS~\cite{zhu2025fsgs}, which makes the optimized model a strong baseline. Based on this we conduct experiments to verify the effectiveness of our method. The model is denoted as `Baseline 3DGS' in the following. 

\noindent \textbf{Implementation Details. }The baseline model described above serves as the model $\mathcal R$ for scene-grounding guidance and trajectory initialization.  
For sequence generation, we employ the camera-controlled image-to-video diffusion model~\cite{yu2024viewcrafter} which supports the generation of $L=25$ frames. The weighting factors, $\lambda$, $\lambda_{\rm{perc}}$, $\lambda_{\rm{gen1}}$, and $\lambda_{\rm{gen2}}$ are set to 0.2, 10$^{-4}$, 0.1, and 0.01, respectively. 
The threshold of $\eta_{\rm{mask}}$ is set to 0.9, while $\eta$ is set to 0.5. 
The generation interval $N_{\rm{gen}}$ is set to 260 and $N_{\rm{iter}}$ is set to 10,000. 
% the random ratio

\begin{figure}[t]
  \centering
  \includegraphics[width=0.99\linewidth]{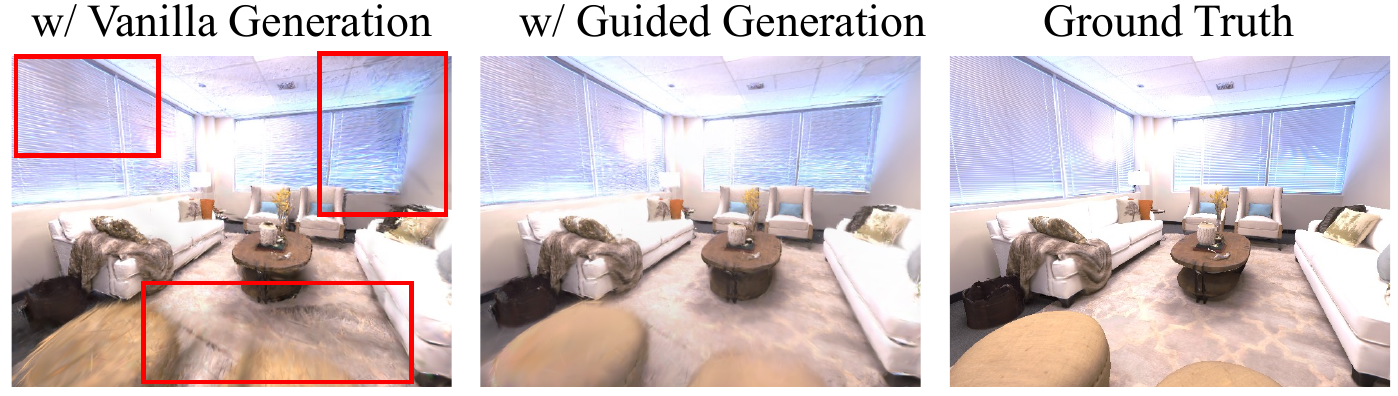}
  \vspace{-3mm}
  \caption{Sequences from the vanilla generation suffer from inconsistencies.~A 3DGS model optimized with these sequences renders images with black shadows, highlighted by red boxes, while our method solves this issue with the scene-grounding guidance. }\vspace{-5mm}
  \label{fig:consistency}
\end{figure}

\begin{figure*}[t]
  \centering
  \includegraphics[width=0.94\linewidth]{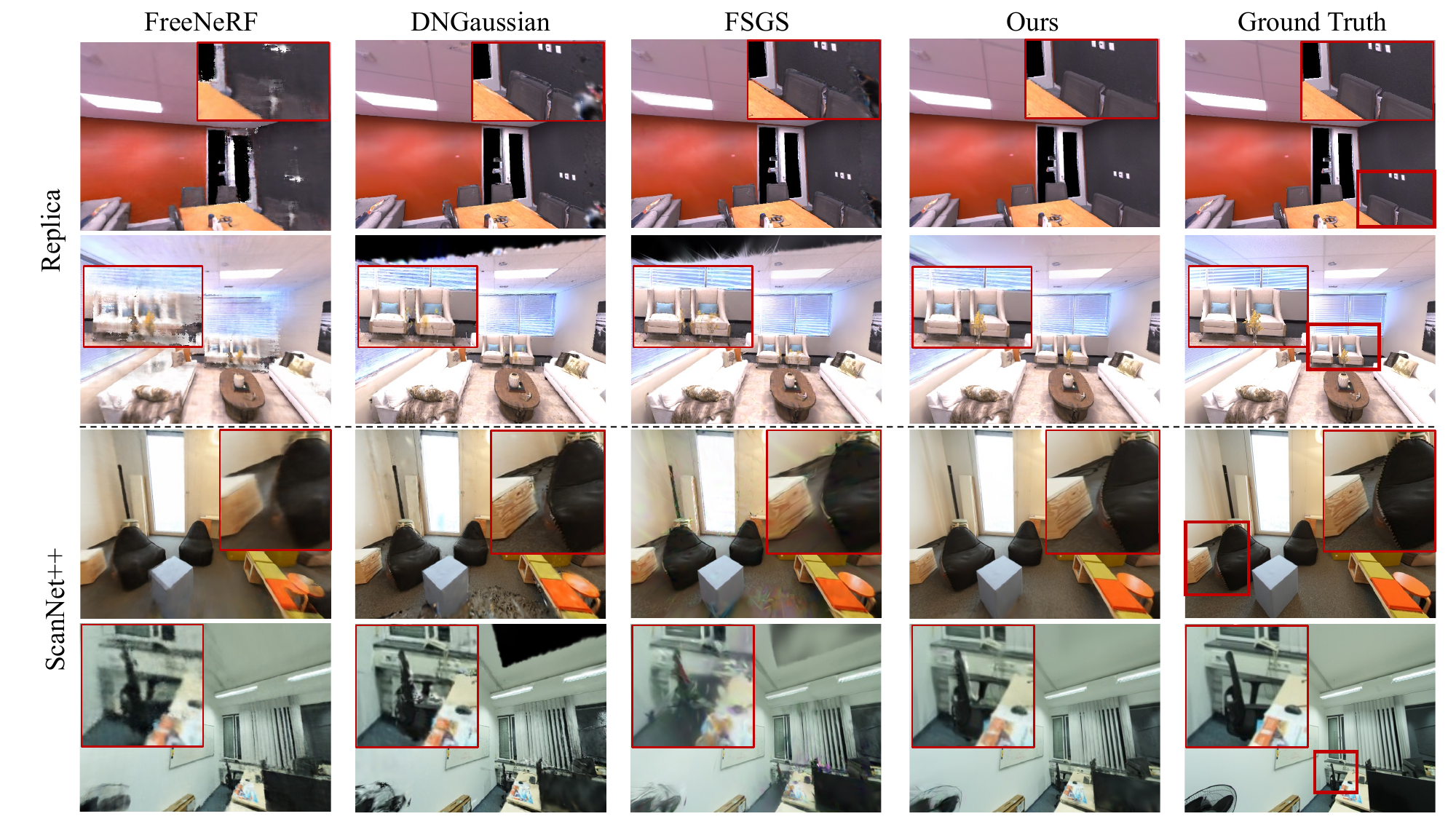}
  \vspace{-3mm}
  \caption{Qualitative comparisons on the Replica and ScanNet++ datasets. All 3DGS-based methods are optimized using the initialized point cloud from DUSt3R~\cite{wang2024dust3r}. 
  Our method effectively addresses the issues of extrapolation and occlusion while preserving finer details and reducing artifacts. For better visualization, please zoom in on the results. }\vspace{-0mm}
  \vspace{-4mm}
  \label{fig:sota1}
\end{figure*}

\subsection{Comparisons}
% For a fair comparison with existing 3DGS-based methods, we evaluate their performances by initializing them with point clouds from DUSt3R~\cite{wang2024dust3r}. 
\noindent \textbf{Comparison on Replica. }
As shown in Tab.~\ref{tab:sota}, our method achieves the highest performance on the Replica dataset, outperforming DNGaussian~\cite{li2024dngaussian} and FSGS~\cite{zhu2025fsgs} by a significant margin of over 3.0 dB in PSNR. 
Fig.~\ref{fig:sota1} illustrates that our method effectively addresses occlusion and extrapolation, while other 3DGS-based methods struggle with these challenges. Additionally, their depth regularization often compromise thin structures, such as the flower in the vase in the second row.
FreeNeRF~\cite{yang2023freenerf} exhibits severe artifacts because it cannot effectively utilize the strong prior from the DUSt3R point cloud. 
Although FreeNeRF can fill hole regions through neighboring interpolation (e.g., the wall behind the chair in the first row and the ceiling in the second row), the results frequently exhibit blurring or artifacts. 

\begin{table}[t]
\centering
\Large
\resizebox{1.\linewidth}{!}{
\renewcommand{\arraystretch}{1.05}
\begin{tabular}{l|ccc|ccc}
\toprule[2pt]
\multicolumn{1}{c|}{\multirow{2}{*}{Method}} & \multicolumn{3}{c|}{Replica~\cite{straub2019replica}} & \multicolumn{3}{c}{ScanNet++~\cite{yeshwanth2023scannet++}} \\
\multicolumn{1}{c|}{}                         & PSNR$\uparrow$     & SSIM$\uparrow$   & LPIPS$\downarrow$    & PSNR$\uparrow$    & SSIM$\uparrow$    & LPIPS$\downarrow$   \\ \midrule
Mip-NeRF~\cite{barron2021mip} & 18.12        & 0.707        & 0.391 & 19.58         &  0.755        &  0.389              \\
InfoNeRF~\cite{kim2022infonerf}     & 13.07        & 0.598     & 0.552   & 14.54         &  0.646        &  0.495             \\
DietNeRF~\cite{jain2021putting}       & 18.99       &0.676     &0.444   & 19.76         & 0.719       & 0.431       \\
FreeNeRF~\cite{yang2023freenerf}   &20.99         &0.765    &0.324     &20.17   &0.756          & 0.368          \\
S$^3$NeRF~\cite{zhong2025empowering} &  22.54   & 0.800  & 0.287 &\cellcolor{orange!25}22.21 & 0.787  &0.364 \\
\midrule[1.2pt]
3DGS$^{\updownarrow}$~\cite{kerbl20233d} &\cellcolor{yellow!25}22.80 & 0.818 & \cellcolor{orange!25}0.179 & \cellcolor{yellow!25}21.41 & \cellcolor{orange!25}0.817  & \cellcolor{orange!25}0.211 \\
DNGaussian~\cite{li2024dngaussian} & 17.63 & 0.718  & 0.435 & 19.01 & 0.754 & 0.367 \\
DNGaussian$^{\updownarrow}$~\cite{li2024dngaussian} & 22.71 & \cellcolor{yellow!25}0.821  & \cellcolor{yellow!25}0.189 & 20.68 & 0.788 & 0.281 \\
FSGS~\cite{zhu2025fsgs} & 20.22 & 0.760 & 0.304 & 17.95 & 0.730 & 0.373 \\
FSGS$^{\updownarrow}$~\cite{zhu2025fsgs} & \cellcolor{orange!25}22.99 & \cellcolor{orange!25}0.833 & 0.205 & 21.23 & \cellcolor{yellow!25}0.813 & \cellcolor{yellow!25}0.257 \\
\midrule[1.2pt]
Ours     & \cellcolor{red!25}{26.35}   & \cellcolor{red!25}{0.872}   &  \cellcolor{red!25}{0.122}   & \cellcolor{red!25}{23.89}     &  \cellcolor{red!25}{0.850}   &  \cellcolor{red!25}{0.182}     \\ 
\bottomrule[2pt]
\end{tabular}}
\vspace{-3mm}
\caption{Quantitative comparisons on the Replica and ScanNet++ datasets. Including our approach, 3DGS-based methods marked with ${\updownarrow}$ are initialized with the point cloud from DUSt3R~\cite{wang2024dust3r}. 
}\vspace{-5.8mm}\label{tab:sota}
\end{table}

\noindent \textbf{Comparison on ScanNet++. }ScanNet++ is a dataset~captured in realistic scenes, so it is more complicated and challenging than the synthetic Replica dataset. 
The results in Tab.~\ref{tab:sota} demonstrate that our method has a clear advantage over current approaches, surpassing FSGS by more than 2.5 dB in PSNR.
As depicted in Fig.~\ref{fig:sota1}, 
our method effectively addresses the extrapolation issue (e.g., the ceiling in the fourth row) and mitigates needle-like artifacts observed in the rendered images of DNGaussian~\cite{li2024dngaussian} and FSGS~\cite{zhu2025fsgs} (the third row).
% in addition to effectively addressing extrapolation issues (e.g., the ceiling in the fourth row), our method reduces needle-like artifacts found in the rendered images of DNGaussian and FSGS (the third row). 
Furthermore, the comparisons in the third row highlight our method's superiority in preserving finer details compared to all other methods. 
\vspace{-1mm}

\begin{figure}[t]
\vspace{-3mm}
  \centering  \includegraphics[width=1\linewidth]{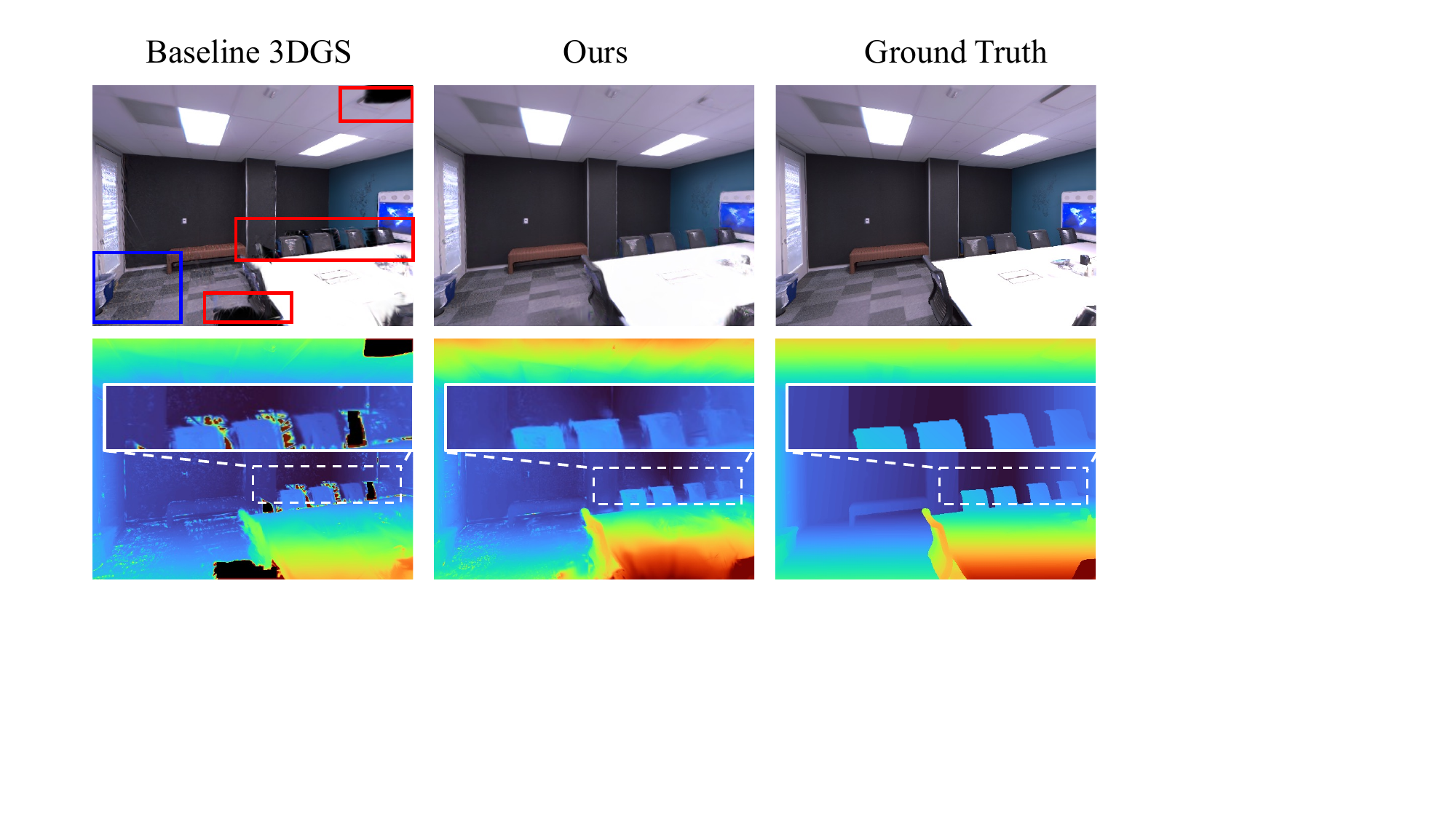}
  \vspace{-7mm}
  \caption{Our method not only effectively addresses extrapolation and occlusion (red boxes), improving the overall quality (blue boxes), but also predicts more plausible geometry. }\vspace{-6mm}
  \label{fig:finalfull}
\end{figure}

\begin{table*}[t]
	\centering
        \Large
	\hspace*{-5mm}
	\makebox[0.59\textwidth][l]{
		\begin{subtable}{0.59\textwidth}
			% \raggedright
			\centering
			\adjustbox{width=1.09\linewidth}{%
                    % \resizebox{}{}{}
                    \renewcommand{\arraystretch}{1.18}
				\begin{tabular}{l|ccc|ccc|ccc}
					\toprule[2pt]
					\multirow{2}{*}{\textbf{(a)}}& \multirow{2}{*}{\textit{Gen.}} & \multirow{2}{*}{\textit{Guide.}} & \multirow{2}{*}{\textit{Traj.}} & \multicolumn{3}{c|}{Full Image} & \multicolumn{3}{c}{Observable Regions} \\
					&                      &                        &                       & PSNR$\uparrow$     & SSIM$\uparrow$     & LPIPS$\downarrow$     & PSNR$\uparrow$    & SSIM$\uparrow$    & LPIPS$\downarrow$   \\ 
					\midrule[1.2pt]
					Baseline 3DGS &    &       &      & 22.80      & 0.818     & 0.179 &25.45&0.860&0.129        \\
					w/ Vanilla Generation         & \checkmark    &       &   & 23.69     & 0.840     & 0.160 &25.00&0.870&0.119         \\
					w/ Guided Generation         &  \checkmark   & \checkmark  &   & 25.03     & 0.852     & 0.139 &26.52&0.881&0.101       \\
					\cellcolor{gray!25}w/ Guided Generation\&Traj.        &  \cellcolor{gray!25}\checkmark   & \cellcolor{gray!25}\checkmark  & \cellcolor{gray!25}\checkmark & \cellcolor{gray!25}\textbf{25.58}     & \cellcolor{gray!25}\textbf{0.859}     & \cellcolor{gray!25}\textbf{0.138}     &\cellcolor{gray!25}\textbf{26.53}&\cellcolor{gray!25}\textbf{0.883}&\cellcolor{gray!25}\textbf{0.100}       \\ 
					\bottomrule[2pt]
				\end{tabular}
			}
			% \caption{Effectiveness of the scene-grounding guidance ($\textit{Guide.}$) and the trajectory initialization ($\textit{Traj.}$). $\textit{Gen.}$ represents optimizing a 3DGS with generated sequences. }
		\end{subtable}
	}
	% \hfill
	\hspace{0.05\textwidth} % Adjust the horizontal space between tables
	\makebox[0.312\textwidth][r]{
		\begin{subtable}{0.311\textwidth}
			\centering
			\adjustbox{width=1.12\linewidth}{%
                    \renewcommand{\arraystretch}{1.18}
				\begin{tabular}{l|ccc}
					\toprule[2pt]
					\textbf{(b)} & PSNR$\uparrow$ & SSIM$\uparrow$ & LPIPS$\downarrow$ \\ 
					\midrule[1.2pt]
				    Baseline 3DGS& 22.80 & 0.818 & 0.179 \\
                        w/ Guided Generation\&Traj. & 25.58     & 0.859     & 0.138      \\
					% w/ Guided Generation$^*$\&Traj. & 25.83     & 0.860     & 0.133     \\
					% w/ perceptual loss     & 26.07     & 0.870     & 0.127      \\ 
					\cellcolor{gray!25}w/ perceptual loss &\cellcolor{gray!25}\textbf{26.35} & \cellcolor{gray!25}\textbf{0.872} & \cellcolor{gray!25}\textbf{0.122} \\
					\quad w/o local sampling                     & 26.28     & 0.871     & 0.127      \\ 
					\quad w/o global list    &26.01      &0.867      & 0.122       \\
					\bottomrule[2pt]
				\end{tabular}
			}
			% \caption{Effectiveness of the proposed strategies for 3DGS optimization. }
	\end{subtable}}
	\vspace{-3mm}\caption{Ablation experiments on the Replica dataset. \textbf{(a)} Effectiveness of the proposed scene-grounding guidance (\textit{Guide.}) for generation, and the trajectory initialization strategy (\textit{Traj.}). (\textit{Gen.}) indicates utilizing generated sequences for modeling. Metrics of observable regions mask out regions outside the field of view or occluded. \textbf{(b)} Effectiveness of the proposed scheme for 3DGS optimization. }
    % $^*$ refers to guiding the generation with an improved baseline. }
	\vspace{-3mm}
	\label{tab:ablation}
\end{table*}

\subsection{Ablation Studies}
\vspace{-1mm}
Our technical contributions consist of three key components. We analyze their effects on the Replica dataset.

\noindent \textbf{Generation with Scene-Grounding Guidance. }
Optimizing a 3DGS with sequences from vanilla generation results in quality degradation. In Tab.~\ref{tab:ablation} (a), while the full image metrics are enhanced due to slightly improved modeling at occluded regions, the visual quality degrades, as indicated by PSNR of observable regions dropping from 25.45 dB to 25.00 dB. This degradation is attributed to inconsistencies within generated sequences, which can result in black shadows in rendered images as illustrated in Fig.~\ref{fig:consistency}.
In contrast, our scene-grounding guidance ensures that the generated sequences remain consistent, significantly enhancing the modeling capability in regions outside the field of view and occluded, while also improving the overall quality, evidenced by the `w/ Guided Generation' results in Tab.~\ref{tab:ablation} (a). 

\begin{figure}[t]
  \centering
  \includegraphics[width=0.99\linewidth]{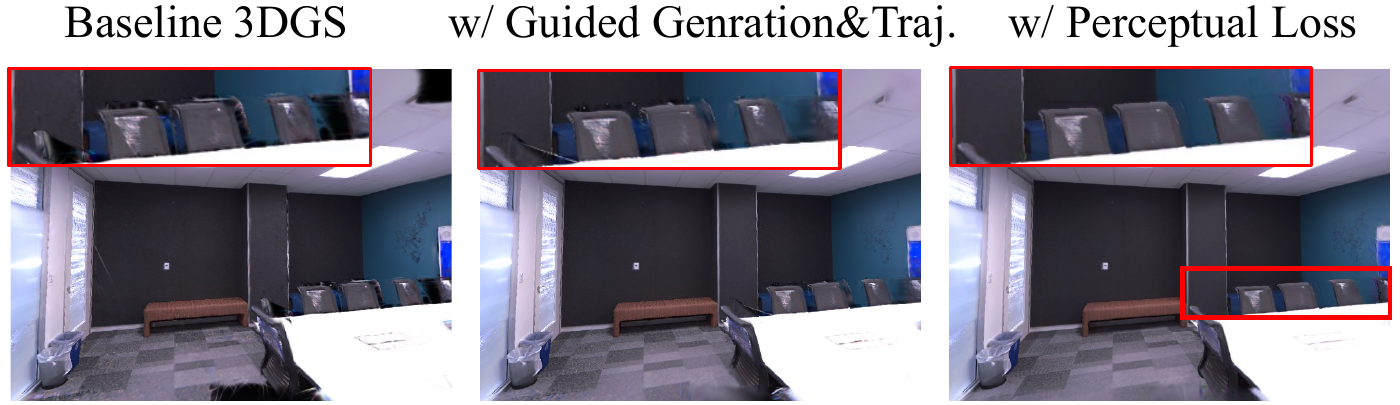}
  \vspace{-3mm}
  \caption{The perceptual loss for generated views greatly increases the modeling capability at hole regions. }\vspace{-3mm}
  \label{fig:percep}
\end{figure}

\begin{table}[t]
    \centering
    \Large
    \adjustbox{width=0.9\linewidth}{%
        \begin{tabular}{l|ccc}
            \toprule[2pt]
            & PSNR$\uparrow$ & SSIM$\uparrow$ & LPIPS$\downarrow$ \\ 
            \midrule[1.2pt]
            Baseline 3DGS+LaMa~\cite{suvorov2022resolution}&   24.56   & 0.833 & 0.167     \\
            Baseline 3DGS+SDS~\cite{poole2022dreamfusion} with SDInpaint &24.33 &0.844 &0.153 \\
            Baseline 3DGS+SDS~\cite{poole2022dreamfusion} with SDInpaint$^*$ &25.15 &0.853 &0.141 \\
            \cellcolor{gray!25}Ours & \cellcolor{gray!25}\textbf{26.35} & \cellcolor{gray!25}\textbf{0.872} & \cellcolor{gray!25}\textbf{0.122} \\ 
            \bottomrule[2pt]
            \end{tabular}
    }\vspace{-3mm}
\caption{Comparisons with inpainting methods on the Replica dataset. $^*$ indicates the usage of our trajectory initialization. }\vspace{-4mm}\label{tab:inpaint}
\end{table}

\noindent \textbf{Trajectory Initialization Strategy.} Tab.~\ref{tab:ablation} (a) further demonstrates that the proposed trajectory initialization strategy significantly boosts the performance, notated as `w/ Guided Generation\&Traj'. The improvement mainly arises from enhanced modeling of the regions outside the field of view or the occluded areas, as the metrics of visible regions plateau while the overall image metrics improve by over 0.5 dB in PSNR. This indicates that the initialization strategy effectively identifies hole regions for holistic modeling. 

\noindent \textbf{Scheme for 3DGS Optimization with Generation. }We verify the effectiveness of the proposed scheme in Tab.~\ref{tab:ablation} (b).
% for 3DGS optimization with generation in Tab.~\ref{tab:ablation} (b). 
% `Guided Generation$^*$\&Traj' guides the generation with an improved baseline 3DGS, detailed in the supplement. 
% We ablate the proposed strategies on our best-performing model to further consolidate their importance. 
Specifically, the perceptual loss of Eq.~\eqref{eq:loss_gen} increases PSNR by over 0.5 dB, which is crucial for the model to fill the hole regions, as shown in Fig.~\ref{fig:percep}. We empirically find that local sampling brings improvement in Sec.~\ref{sec:train_gs}. This is evidenced by the performance decrease of `w/o local sampling', which randomly samples generated views from all generated sequences. 
Alg.~\ref{alg:optim} shows that we use a global list to avoid the forgetting problem, and its necessity is verified by an over 0.3 dB PSNR drop observed with `w/o global list'. 
Combining these contributions, our full model effectively addresses extrapolation and occlusion while enhancing overall image quality, meanwhile exhibiting much better geometry, as shown in Fig.~\ref{fig:finalfull}.

\begin{figure}[t]
  \centering
  \includegraphics[width=0.99\linewidth]{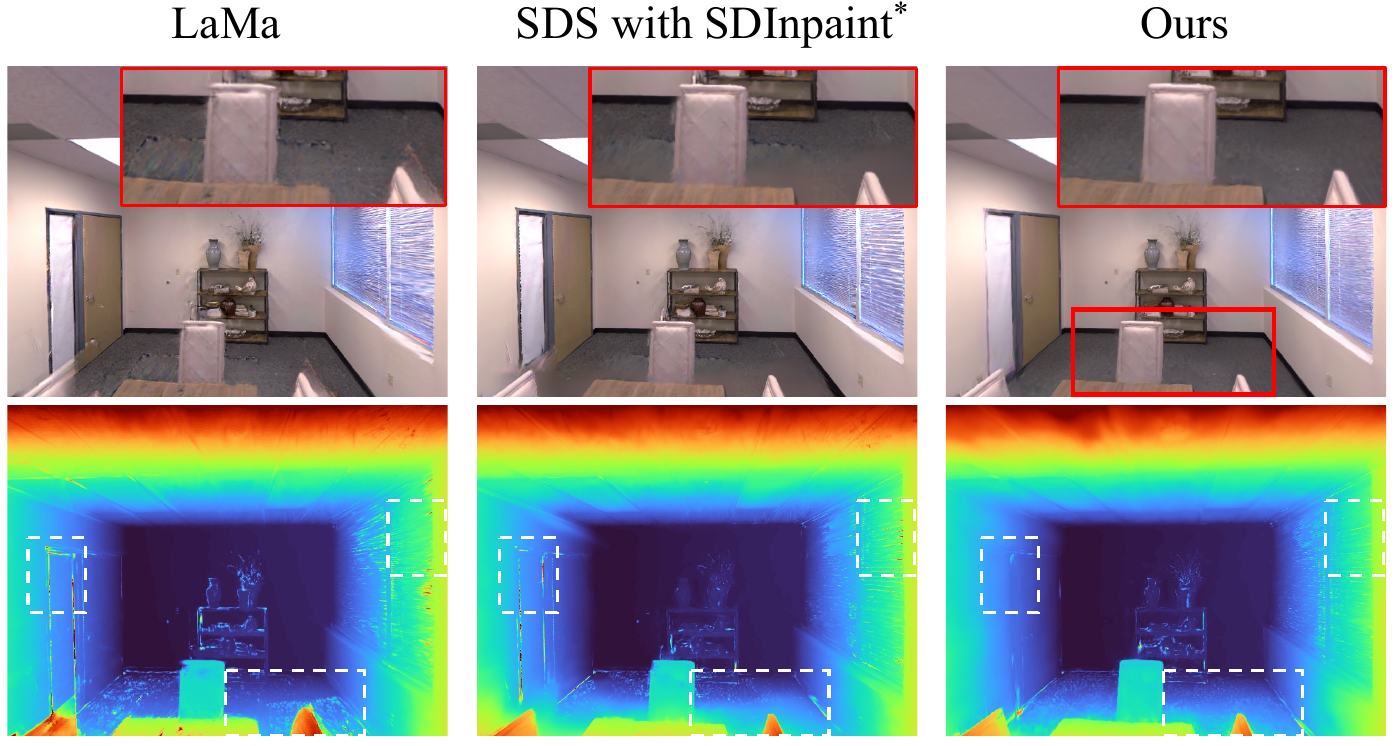}
  \vspace{-3mm}
  \caption{Qualitative comparisons with other inpainting methods. Our approach not only produces more plausible appearances around the inpainting regions but also predicts more consistent geometries in fine-grained local areas. }
  \vspace{-5mm}
  \label{fig:inpaint}
\end{figure}

\subsection{Further Comparisons with Inpainting Methods}
\vspace{-1mm}
% \noindent \textbf{Comparison with Inpainting Methods. }
Extrapolation and occlusion can also be addressed using inpainting methods. We thus compare our approach with two inpainting-based methods. One method applies LaMa~\cite{suvorov2022resolution} inpainting on hole regions, while the other optimizes a 3DGS by Score Distillation Sampling (SDS)~\cite{poole2022dreamfusion} based on a SDInpaint model~\cite{rombach2022high}. We also incorporate our proposed trajectory initialization into the SDS method to enhance the optimization of inpainting regions. Results in Tab.~\ref{tab:inpaint} show that our method outperforms these two methods by more than 1.0 dB in PSNR. Qualitative results in Fig.~\ref{fig:inpaint} indicate that, under certain conditions, the SDS-based method produces inpainted areas with strange appearances, while LaMa tends to create blurring artifacts interpolated from neighboring regions. 
Besides, Fig.~\ref{fig:inpaint} also shows that our method predicts better geometry in regions with local details. 
The inpainting results from our approach are more plausible due to the well-designed guidance, which effectively exploits prior knowledge from the diffusion model. 
 
% \noindent \textbf{Guidance Visualization. }

\section{Conclusion}
\vspace{-1mm}
In this paper, we have explored to address the critical issues of extrapolation and occlusion in sparse-input 3DGS modeling. We propose using video diffusion models that provide plausible interpretations for regions that are outside the field of view and occluded. To resolve inconsistencies within generated sequences, we introduce a novel scene-grounding guidance that controls the diffusion model to generate consistent sequences without any fine-tuning. Additionally, we propose a trajectory initialization strategy to enhance holistic modeling and develop a scheme for optimizing 3DGS with generated sequences.~Extensive experiments validate our approach, demonstrating that it outperforms current methods by a significant margin. 

% \FloatBarrier
% \newpage 
{
    \small
    \bibliographystyle{ieeenat_fullname}
    \bibliography{main}
}


\clearpage
\setcounter{page}{1}
\maketitlesupplementary

\renewcommand{\thefigure}{A\arabic{figure}}
\renewcommand{\thetable}{A\arabic{table}}
\renewcommand{\thesection}{\Alph{section}}
\setcounter{figure}{0}
\setcounter{table}{0}
\setcounter{section}{0}

% \textit{We recommend reviewers to check the HTML page in the supplementary material (please click $\textbf{\rm{guidegs/demo.html}}$), which contains demo videos for better visualization. }

\section{Implementation Details}
\vspace{-2mm}
During the denoising sampling process, we employ the DDIM sampler~\cite{song2020denoising} combined with our proposed guidance, setting the number of sampling steps to 50. 
Regarding the trajectory initialization strategy, for each input view in its camera space, we sample views by changing the polar/azimuth angle to { $[-30^{\circ},-15^{\circ},0^{\circ},15^{\circ},30^{\circ}]$}, and setting the radial distance to { $[1, \frac{1}{3}, \frac{1}{10}]$} of the depth of the center pixel (from the prediction of ViewCrafter~\cite{yu2024viewcrafter}).
Out of 75 sampled views, we discard those whose renderings exhibit holes larger than 10$\%$ of the image size (to filter out uncommon viewpoints), then select the top 6 views with the largest holes from the remaining. 
To obtain the point cloud used for initialization, we follow the standard pipeline provided on the DUSt3R~\cite{wang2024dust3r} webpage. Since our focus is sparse-input radiance fields reconstruction, the groundtruth camera poses and intrinsics are provided. During DUSt3R optimization, we fix both the poses and intrinsics to their groundtruth values.

\section{More Results}
\vspace{-2mm}

Our method focuses on holistic modeling of an indoor scene of a moderate size, and we conduct the experiments in the main paper with 6 input views, since 6 input views are basically sufficient to cover the entire room. To validate the effectiveness of our method, we also test our method with different number of views following the common 3/6/9-view settings of sparse-input modeling. Tab.~\ref{tab:suppl_view} validates that, our method is effective given different number of input views, with consistent improvements over our baseline. InstantSplat~\cite{fan2024instantsplat} is a strong baseline of sparse-input pose-free modeling, leveraging DUSt3R~\cite{wang2024dust3r} point cloud for 3DGS initialization. 
Our method also consistently outperforms InstantSplat as shown in Tab.~\ref{tab:suppl_view}. 

To obtain a thorough understanding of the source of the performance improvement, we show some quantitative results regarding performances of observable and the unobservable regions respectively in Tab.~\ref{tab:suppl_obs}. The results show that our method brings improvement in both observable and unobservable regions.

We further compare our method with two representative methods that leverage diffusion models for sparse-input modeling, ReconFusion~\cite{wu2024reconfusion} and CAT3D~\cite{gao2024cat3d} on the datasets of RealEstate10K and LLFF. We adhere to their settings for fair comparisons and the results are shown in Tab.~\ref{tab:suppl_cat3d}. On the LLFF dataset, our method is based on the strong baseline of binocular-guided 3DGS~\cite{han2025binocular}. The results show that our method achieves comparable performance with both ReconFusion and CAT3D.

\begin{figure*}[th]
  \centering
  \includegraphics[width=0.9\linewidth]{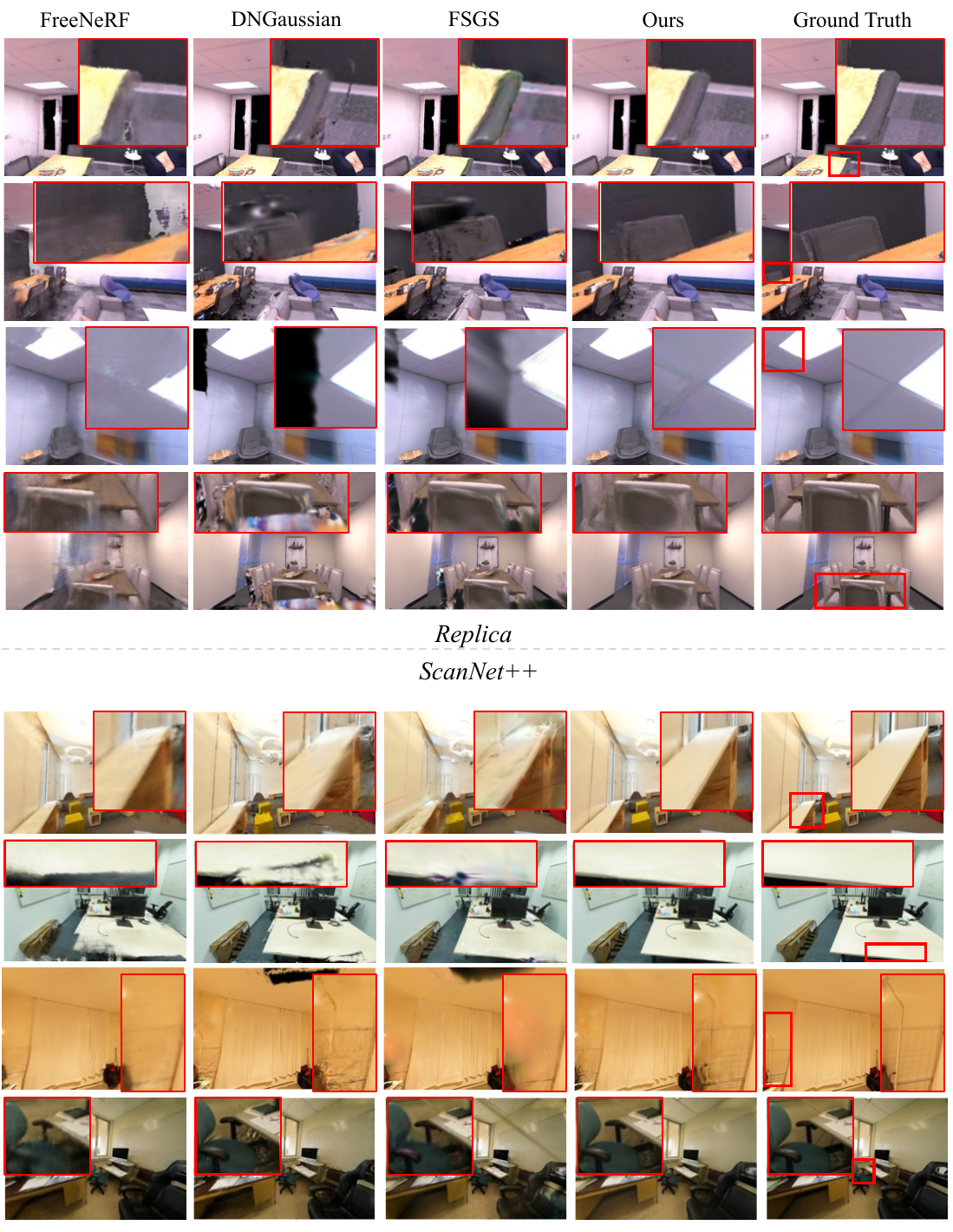}\vspace{-3mm}
  \caption{Qualitative comparisons between other works on Replica and ScanNet++ datasets. All 3DGS-based methods are optimized using the initialized point cloud from DUSt3R~\cite{wang2024dust3r}.  
  }\vspace{5mm}
  \label{fig:suppl_comparison}
\end{figure*}

We provide per-scene comparisons in Table~\ref{tab:suppl_per_scene}, demonstrating that our method consistently achieves superior performance across all scenes. Additional qualitative results are shown in Fig.~\ref{fig:suppl_comparison}. These results highlight the effectiveness of our approach in addressing issues such as extrapolation and occlusion, as seen in examples like the wall behind the chair (second row) and the ceiling (third row). Furthermore, our method preserves more intact structures with finer details, such as the edges in the fifth and sixth rows. 

\begin{figure*}[t]
  \centering
  \vspace{-2mm}
  \includegraphics[width=0.99\linewidth]{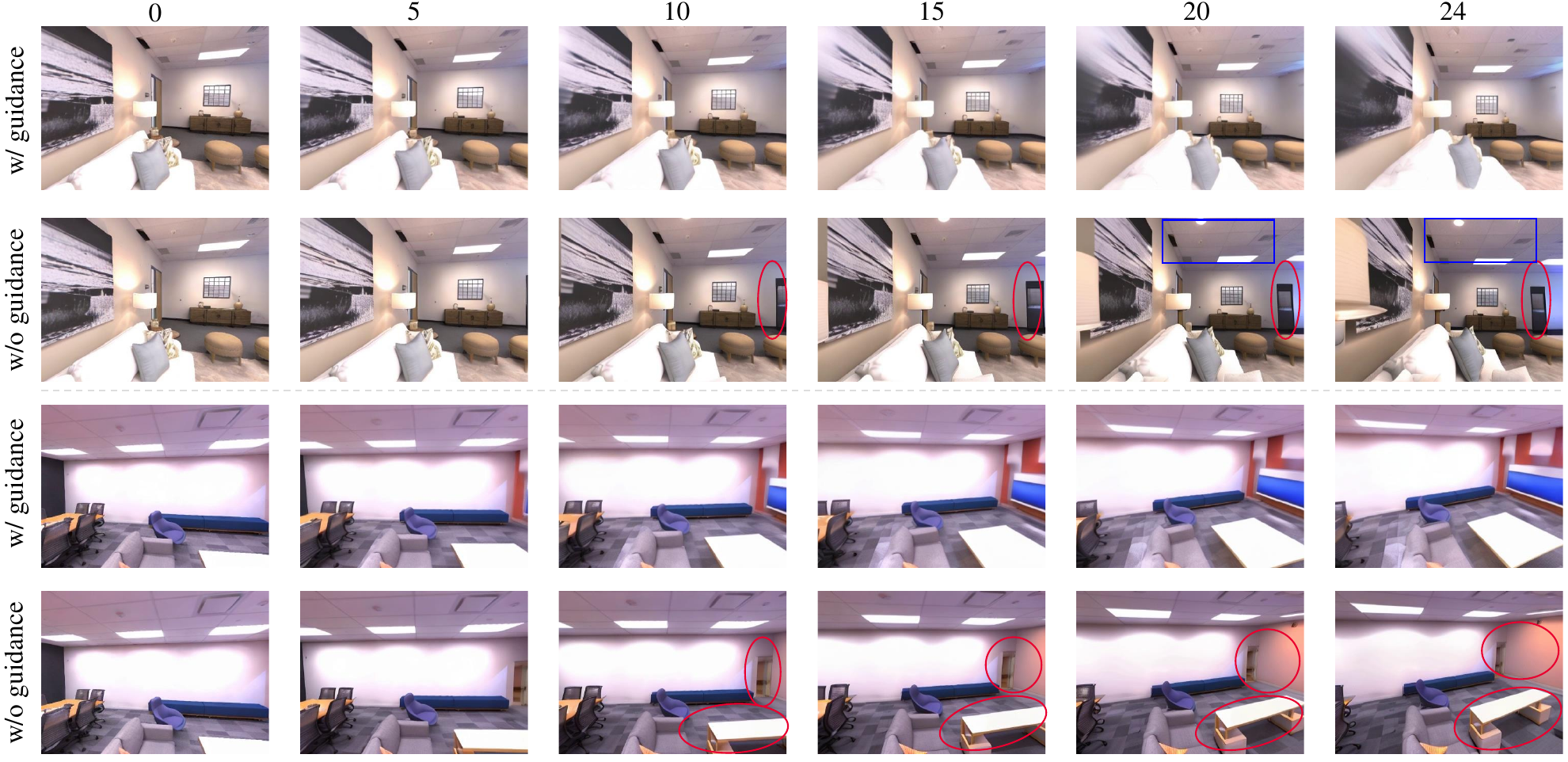}\vspace{-3mm}
  \caption{Generated frames from the video diffusion model with and without the proposed guidance. The numbers at the top indicate the frame IDs. The first frame corresponds to an image from the sparse input views, while other frames are generated. Without guidance, the generated sequences exhibit significant inconsistencies: \textbf{(i)} appearance inconsistencies, highlighted by the blue boxes; and \textbf{(ii)} hallucinated elements that do not exist in the scene, highlighted by the red boxes. In contrast, with the proposed guidance, the generated sequences are more plausible and consistent. 
  }\vspace{-2mm}
  \label{fig:suppl_consistency}
\end{figure*}

We present a comparison of the generated sequences from the video diffusion model with and without the proposed guidance in Fig.~\ref{fig:suppl_consistency}. The results clearly show that our proposed guidance enhances the plausibility of the generated sequences by maintaining consistent appearances and ensuring that only elements present in the scene are generated. Consistency in the generated video is crucial for effective 3DGS optimization. Using inconsistent sequences for 3DGS optimization often leads to artifacts, such as black shadows in the renderings, which significantly degrade visual quality, as demonstrated on the demo page.

\begin{figure*}[t]
  \centering
  \includegraphics[width=0.9\linewidth]{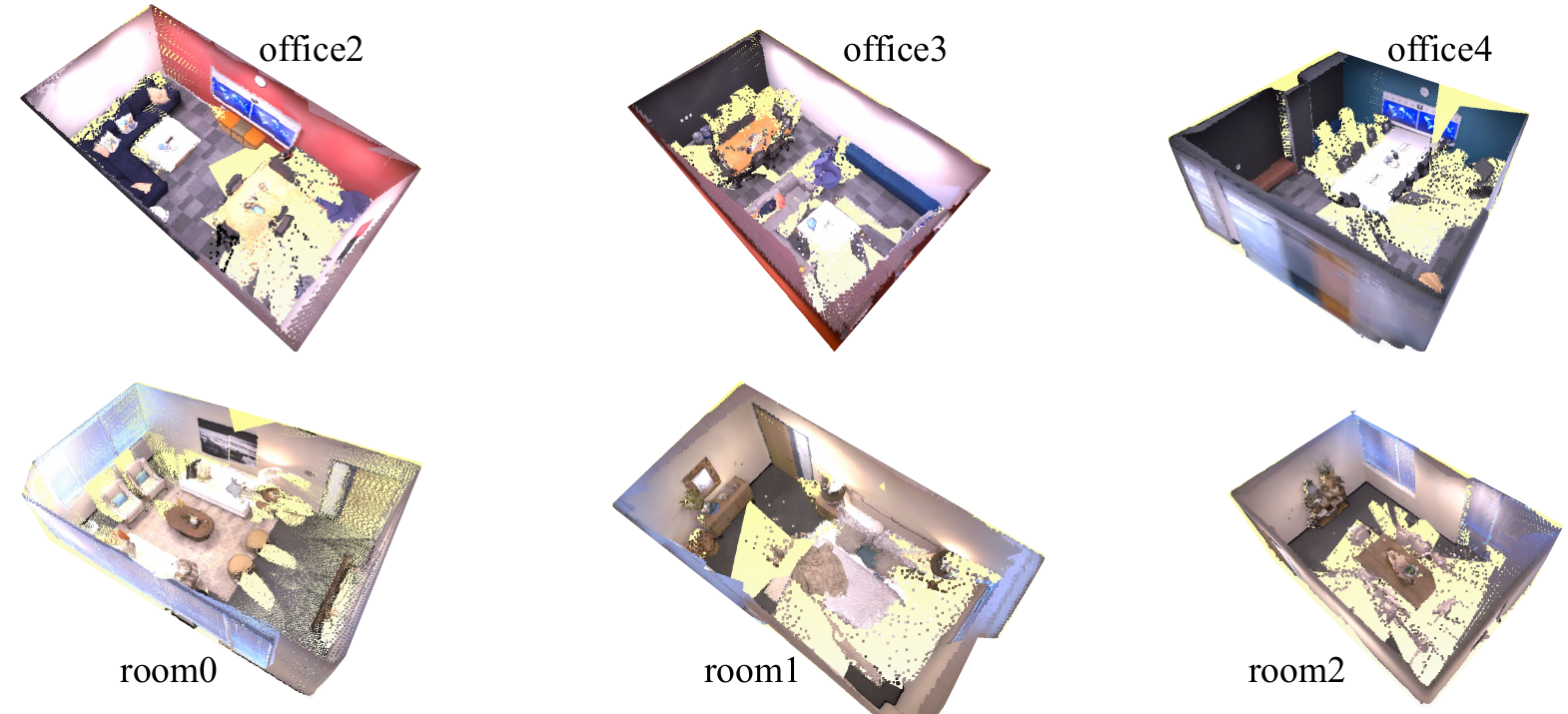}\vspace{-1mm}
  \caption{Point clouds from DUSt3R~\cite{wang2024dust3r} optimized with sparse input views on the Replica dataset. The yellow parts represent unobserved regions, e.g., regions that are outside the field of view or occluded. Note that the ceilings are removed for better visualization. 
  }\vspace{0mm}
  \label{fig:suppl_holes}
\end{figure*}

\begin{table*}[th]
\centering
\resizebox{0.7\linewidth}{!}{
\begin{tabular}{lccccccccc}
\toprule[2pt]
\multicolumn{1}{c|}{\multirow{2}{*}{\textbf{}}} & \multicolumn{3}{c|}{3-view}                & \multicolumn{3}{c|}{6-view}                & \multicolumn{3}{c}{9-view} \\
\multicolumn{1}{c|}{}                           & PSNR$\uparrow$  & SSIM$\uparrow$  & \multicolumn{1}{c|}{LPIPS$\downarrow$} & PSNR$\uparrow$  & SSIM$\uparrow$  & \multicolumn{1}{c|}{LPIPS$\downarrow$} & PSNR$\uparrow$    & SSIM$\uparrow$    & LPIPS$\downarrow$  \\ \midrule[1.5pt]
\multicolumn{1}{l}{\textbf{Replica}}              & & & & &  & & & &   \\
\multicolumn{1}{l|}{Baseline 3DGS}              & 19.87 & 0.794 & \multicolumn{1}{c|}{0.178} & 22.80 & 0.818 & \multicolumn{1}{c|}{0.179} & 24.81   & 0.863   & 0.124  \\
\multicolumn{1}{l|}{InstantSplat~\cite{fan2024instantsplat}}               & 20.49 & 0.766 & \multicolumn{1}{c|}{0.226} & 20.35 & 0.760 & \multicolumn{1}{c|}{0.290} & 18.44   & 0.708   & 0.373  \\
\multicolumn{1}{l|}{Ours}                       & \cellcolor{orange!60}23.98 & \cellcolor{orange!60}0.848 & \multicolumn{1}{c|}{\cellcolor{orange!60}0.136} & \cellcolor{orange!60}26.35 & \cellcolor{orange!60}0.872 & \multicolumn{1}{c|}{\cellcolor{orange!60}0.122} & \cellcolor{orange!60}27.42   & \cellcolor{orange!60}0.891   & \cellcolor{orange!60}0.111  \\ 
\bottomrule[2pt]
\end{tabular}
}\vspace{-1.5mm}
\caption{
Our method brings performance improvement over the baseline with different number of input views, and consistently outperforms another strong sparse-input modeling baseline InstantSplat~\cite{fan2024instantsplat}. }\label{tab:suppl_view}
%Baseline 3DGS is initialized with DUSt3R point cloud. 
\end{table*}

\begin{table*}[th]
\centering
\resizebox{0.8\linewidth}{!}{
\begin{tabular}{lccccccccc}
\toprule[2pt]
\multicolumn{1}{c|}{\multirow{2}{*}{\textbf{Replica} 6-view}} & \multicolumn{3}{c|}{Full Image}                & \multicolumn{3}{c|}{Observable Regions}                & \multicolumn{3}{c}{Unobservable Regions} \\
\multicolumn{1}{c|}{}                           & PSNR$\uparrow$  & SSIM$\uparrow$  & \multicolumn{1}{c|}{LPIPS$\downarrow$} & PSNR$\uparrow$  & SSIM$\uparrow$  & \multicolumn{1}{c|}{LPIPS$\downarrow$} & PSNR$\uparrow$    & SSIM$\uparrow$    & LPIPS$\downarrow$  \\ \midrule[1.5pt]
\multicolumn{1}{l|}{Baseline 3DGS}              & 22.80 & 0.818 & \multicolumn{1}{c|}{0.179} & 25.45 & 0.860 & \multicolumn{1}{c|}{0.129} & 14.27   & 0.967   & 0.029  \\
\multicolumn{1}{l|}{w/ Vanilla Generation}               & 23.69 & 0.840 & \multicolumn{1}{c|}{0.160} & 25.00 & 0.870 & \multicolumn{1}{c|}{0.119} & 17.11   & 0.977   & 0.025  \\
\multicolumn{1}{l|}{Ours}                       & \cellcolor{orange!60}26.35 & \cellcolor{orange!60}0.872 & \multicolumn{1}{c|}{\cellcolor{orange!60}0.122} & \cellcolor{orange!60}27.12 & \cellcolor{orange!60}0.894 & \multicolumn{1}{c|}{\cellcolor{orange!60}0.091} & \cellcolor{orange!60}20.85   & \cellcolor{orange!60}0.985   & \cellcolor{orange!60}0.020                              \\ \midrule[1.5pt]
\multicolumn{1}{l|}{Baseline 3DGS+LaMa~\cite{suvorov2022resolution}} &24.56 &0.833 &\multicolumn{1}{c|}{0.167} &25.45 &0.860 &\multicolumn{1}{c|}{0.129} &17.80 &0.981 &0.021 \\
\multicolumn{1}{l|}{Baseline 3DGS+SDInpaint~\cite{rombach2022high}$^*$} &25.15 &0.853 &\multicolumn{1}{c|}{0.141} &26.13 &0.878 &\multicolumn{1}{c|}{0.104} &19.25 &0.982 & 0.022 \\  \bottomrule[2pt]
\end{tabular}
}\vspace{-1.5mm}
\caption{{Analysis of performance regarding observable and unobservable regions. $^*$ refers to incorporating our trajectory initialization strategy. The methods in the second block utilize inpainting models. }}\label{tab:suppl_obs}
\end{table*}

\begin{table*}[t]
\centering
\resizebox{0.75\linewidth}{!}{
\begin{tabular}{lccccccccc}
\toprule[2pt]
\multicolumn{1}{c|}{\multirow{2}{*}{\textbf{}}} & \multicolumn{3}{c|}{3-view}                & \multicolumn{3}{c|}{6-view}                & \multicolumn{3}{c}{9-view} \\
\multicolumn{1}{c|}{}                           & PSNR$\uparrow$  & SSIM$\uparrow$  & \multicolumn{1}{c|}{LPIPS$\downarrow$} & PSNR$\uparrow$  & SSIM$\uparrow$  & \multicolumn{1}{c|}{LPIPS$\downarrow$} & PSNR$\uparrow$    & SSIM$\uparrow$    & LPIPS$\downarrow$  \\ \midrule[1.5pt]
% \multicolumn{10}{c}{\textbf{RealEstate10K}}                                                                                                                                           \\ \midrule[1.5pt]
\multicolumn{1}{l}{\textbf{RealEstate10K}}              & & & & &  & & & &   \\
\multicolumn{1}{l|}{ReconFusion~\cite{wu2024reconfusion}}              & \cellcolor{orange!20}25.84 & \cellcolor{orange!20}0.910 & \multicolumn{1}{c|}{0.144} & 29.99 & \cellcolor{orange!20}0.951 & \multicolumn{1}{c|}{0.103} & 31.82   &0\cellcolor{orange!20}.961   & 0.092  \\
\multicolumn{1}{l|}{CAT3D~\cite{gao2024cat3d}}               & \cellcolor{orange!60}26.78 & \cellcolor{orange!60}0.917 & \multicolumn{1}{c|}{\cellcolor{orange!60}0.132} & \cellcolor{orange!60}31.07 & \cellcolor{orange!60}0.954 & \multicolumn{1}{c|}{\cellcolor{orange!20}0.092} & \cellcolor{orange!20}32.20   & \cellcolor{orange!60}0.963   & \cellcolor{orange!20}0.082  \\
\multicolumn{1}{l|}{Ours} &25.03 & 0.871 & \multicolumn{1}{c|}{\cellcolor{orange!20}0.136} &\cellcolor{orange!20}30.62 & 0.944 & \multicolumn{1}{c|}{\cellcolor{orange!60}0.069}  &\cellcolor{orange!60}32.45  &0.955    &\cellcolor{orange!60}0.062      \\ \midrule[1pt]
\multicolumn{1}{l}{\textbf{LLFF}}              & & & & &  & & & &   \\
\multicolumn{1}{l|}{ReconFusion~\cite{wu2024reconfusion}}              & 21.34 & 0.724 & \multicolumn{1}{c|}{0.203} & 24.25 & 0.815 & \multicolumn{1}{c|}{0.152} & 25.21   & 0.848   & 0.134  \\
\multicolumn{1}{l|}{CAT3D~\cite{gao2024cat3d}}               & \cellcolor{orange!60}21.58     &  \cellcolor{orange!20}0.731     & \multicolumn{1}{c|}{\cellcolor{orange!20}0.181}      & \cellcolor{orange!20}24.71     & \cellcolor{orange!20}0.833      & \multicolumn{1}{c|}{\cellcolor{orange!20}0.121}      & \cellcolor{orange!20}25.63        &  \cellcolor{orange!20}0.860      & \cellcolor{orange!20}0.107       \\
\multicolumn{1}{l|}{Ours}          &\cellcolor{orange!20}21.35       &\cellcolor{orange!60}0.746       & \multicolumn{1}{c|}{\cellcolor{orange!60}0.173}      &\cellcolor{orange!60}25.13       & \cellcolor{orange!60}0.851      & \multicolumn{1}{c|}{\cellcolor{orange!60}0.102}      &  \cellcolor{orange!60}26.29      &\cellcolor{orange!60}0.880         & \cellcolor{orange!60}0.084       \\ 
\bottomrule[2pt]
\end{tabular}
}\vspace{-2mm}
\caption{{Comparisons with ReconFusion~\cite{wu2024reconfusion} and CAT3D~\cite{gao2024cat3d} on the RealEstate10K and LLFF datasets. }}\label{tab:suppl_cat3d}
%Baseline 3DGS is initialized with DUSt3R point cloud. 
\end{table*}

\section{Discussion}
\vspace{-2mm}
While our approach significantly improves overall quality by addressing extrapolation and occlusion challenges, we observe that it occasionally produces over-smoothed results. We hypothesize that this is due to the limited resolution supported by the video diffusion model during generation. On a 32GB V100 GPU, we are constrained to generating sequences at resolutions of 320×448 for the Replica dataset and 320×512 for the ScanNet++ dataset, which are subsequently upsampled to rendering resolutions of 480×640 and 480×720, respectively, for supervision during 3DGS optimization. This upsampling process introduces undersampling, which can smooth out certain regions and result in over-smoothed effects.
Addressing the challenge of preserving high-frequency details during 3DGS optimization under resource-limited sequence generation remains an open problem and is a direction for future work.
% With limited resources for sequence generation, keeping high-frequency details during 3DGS optimization may be left as our future work. 

\begin{table*}[th]
\centering
\scriptsize
\begin{tabular}{c|ccccc|ccccccc}
\toprule
\multirow{2}{*}{}           & \multicolumn{5}{c|}{ScanNet++~\cite{yeshwanth2023scannet++}}                                                     & \multicolumn{7}{c}{Replica~\cite{straub2019replica}}                                                                                          \\
                            & \textit{a2ccc}          & \textit{8a20d}          & \textit{94ee1}          & \textit{78318}          & \textit{avg}          & \textit{office2}        & \textit{office3}        & \textit{office4}        & \textit{room0}          & \textit{room1}          & \textit{room2}          & \textit{avg}           \\ \midrule
   & 18.28          & 23.48          & 16.93          & 19.63          & 19.58          & 17.43          & 19.04          & 19.08          & 17.46          & 16.57          & 19.16          & 18.12          \\Mip-NeRF~\cite{barron2021mip}
                            & 0.759          & 0.799          & 0.725          & 0.735          & 0.755          & 0.539          & 0.685          & 0.727          & 0.762          & 0.721          & 0.808          & 0.707          \\
                    & 0.351          & 0.321          & 0.431          & 0.451          & 0.389          & 0.486          & 0.421          & 0.393          & 0.342          & 0.386          & 0.317          & 0.391          \\ \midrule
 & 13.90          & 17.69          & 14.34          & 12.21          & 14.54          & 13.66          & 12.53          & 11.51          & 12.58          & 14.11          & 14.00          & 13.07          \\InfoNeRF~\cite{kim2022infonerf}
                            & 0.662          & 0.691          & 0.627          & 0.605          & 0.646          & 0.463          & 0.545          & 0.592          & 0.618          & 0.689          & 0.678          & 0.598          \\
     & 0.468          & 0.437          & 0.516          & 0.558          & 0.495          & 0.612          & 0.623          & 0.624          & 0.542          & 0.435          & 0.477          & 0.552          \\ \midrule
   & 20.67          & 23.00          & 15.34          & 20.02          & 19.76          & 19.12          & 19.35          & 18.97          & 19.84          & 17.18          & 19.46          & 18.99          \\DietNeRF~\cite{jain2021putting}
                            & 0.751          & 0.776          & 0.627          & 0.725          & 0.719          & 0.612          & 0.695          & 0.419          & 0.783          & 0.749          & 0.797          & 0.676          \\
  & 0.385          & 0.363          & 0.516          & 0.459          & 0.431          & 0.458          & 0.417          & 0.721          & 0.34           & 0.386          & 0.343          & 0.444          \\ \midrule

   & 19.93          & 22.37          & 19.42          & 18.94          & 20.17          & 20.89          & 21.06          & 20.25          & 22.55          & 19.69          & 21.43          & 20.99          \\FreeNeRF~\cite{yang2023freenerf}
                            & 0.759          & 0.791          & 0.762          & 0.711          & 0.756          & 0.688          & 0.735          & 0.750          & 0.831          & 0.781          & 0.807          & 0.765          \\
                    & {0.307}          & 0.299          & 0.417          & 0.449          & 0.368          & 0.359          & 0.340          & 0.364          & 0.234          & 0.325          & 0.321          & 0.324          \\ \midrule

     & 21.81 & 25.60 & 20.05 & 21.36 & 22.21 & 22.79 & 23.83 & 23.08 & 24.01 & 19.66 & 21.87 & 22.54 \\
                       S$^3$NeRF~\cite{zhong}& 0.801 & 0.811 & 0.784 & 0.753 & 0.787 & 0.728 & 0.773 & 0.801 & 0.862 & 0.808 & 0.825 & 0.800 \\ & 0.324 & 0.330 & 0.357 & 0.444 & 0.364 & 0.326 & 0.309 & 0.301 & 0.213 & 0.277 & 0.293 & 0.287 \\

    \midrule

   & 20.65         & 23.49         & 20.38          & 21.11          & 21.41          & 25.03          & 23.60         & 22.14          & 20.32          & 22.68          & 23.07          & 22.80          \\3DGS$^\updownarrow$~\cite{kerbl20233d}
                            & 0.824          & 0.857          & 0.821          & 0.764          & 0.817          & 0.873          & 0.858         & 0.834          & 0.720          & 0.802          & 0.824          & 0.818          \\
                     & 0.193          & 0.136         & 0.218          & 0.298          & 0.211          & 0.141          & 0.147          & 0.180          & 0.204          & 0.203          & 0.196          & 0.179          \\ \midrule

   & 19.10          & 21.21          & 17.55          & 18.20          & 19.01          & 22.68          & 18.40          & 12.31          & 12.60          & 18.87          & 20.91          & 17.63          \\DNGaussian~\cite{li2024dngaussian}
                            & 0.765          & 0.781          & 0.743          & 0.730          & 0.755          & {0.843}          & 0.789          & 0.644          & 0.534          & 0.708          & 0.790          & 0.718          \\
                     & 0.343          &{0.292}         & 0.382         & 0.450          & 0.367          & {0.233}          & 0.291          &0.628          & 0.722          & 0.397          & 0.338          & 0.435          \\ \midrule

   & 20.47          & 23.73          & 18.90          & 19.61          & 20.68          & 25.31          & 23.34          & 21.83          & 20.33          & 22.59           & 22.88          & 22.71          \\
  DNGaussian$^\updownarrow$~\cite{li2024dngaussian}                          & 0.805          & 0.842          & 0.784          & 0.722          & 0.788          & 0.890          & 0.853          & 0.837          & 0.729          & 0.800          & 0.820          & 0.821          \\
                    & 0.213          & 0.183          & 0.287          & 0.357          & 0.281          & 0.124          & 0.161          & 0.197          & 0.226          & 0.208          & 0.219          & 0.189          \\ 
\midrule

  & 19.19          & 18.98          & 15.77          & 17.87          & 17.95          & 20.70          & 20.26          & 21.62          & 19.65          & 19.23          & 19.89          & 20.22          \\FSGS~\cite{zhu2025fsgs}
                            & 0.760          & 0.735          & 0.719          & 0.708          & 0.730          & 0.802          & {0.790}         & {0.825}          & 0.654          & 0.712          & 0.779          & 0.760          \\
                    & 0.321          & 0.316         & 0.415          & {0.442}          & 0.373          & 0.266          & {0.255}          & {0.271}          & 0.315          & 0.374          & 0.342          & 0.304          \\ \midrule
   &21.28&22.56&20.28&20.79&21.23 & 24.37          & 23.41          & 23.45          & 21.02          & 23.56          & 22.14          & 22.99                    \\FSGS$^\updownarrow$~\cite{zhu2025fsgs}
                         &0.826&0.844&0.815&0.767&0.813   & 0.873          & 0.856          & 0.862          & 0.759          & 0.823          & 0.822          & 0.833          \\
                     & 0.219          & 0.193          & 0.267          & 0.350          & 0.257         & 0.194          & 0.174          & 0.189          & 0.198          & 0.205          & 0.270          & 0.205          \\ 

\midrule
 & \textbf{25.21} & \textbf{25.10} & \textbf{23.10} & \textbf{22.16} & \textbf{23.89} & \textbf{27.46} & \textbf{26.81} & \textbf{27.43} & \textbf{24.85} & \textbf{26.00} & \textbf{25.53} & \textbf{26.35} \\
   Ours                         & \textbf{0.857} & \textbf{0.882} & \textbf{0.860} & \textbf{0.803} & \textbf{0.850} & \textbf{0.916} & \textbf{0.902} & \textbf{0.897} & \textbf{0.796} & \textbf{0.851} & \textbf{0.872} & \textbf{0.872} \\
                     & \textbf{0.157} & \textbf{0.118} & \textbf{0.201} & \textbf{0.269} & \textbf{0.182} & \textbf{0.083} & \textbf{0.099} & \textbf{0.122} & \textbf{0.145} & \textbf{0.142} & \textbf{0.142} & \textbf{0.122} \\ \bottomrule
\end{tabular}\vspace{-2mm}
\caption{Per-scene performance of various models on the ScanNet++ and Replica datasets. For each method, the three rows represent PSNR, SSIM, and LPIPS, respectively. \textit{avg} indicates the average performance across all scenes in each dataset. Including our approach, 3DGS-based methods marked with $^\updownarrow$ are initialized with the point cloud from DUSt3R~\cite{wang2024dust3r}. }\label{tab:suppl_per_scene}\vspace{-3mm}
\end{table*}

% 
% To split the supplementary pages from the main paper, you can use \href{https://support.apple.com/en-ca/guide/preview/prvw11793/mac#:~:text=Delete%20a%20page%20from%20a,or%20choose%20Edit%20%3E%20Delete).}{Preview (on macOS)}, \href{https://www.adobe.com/acrobat/how-to/delete-pages-from-pdf.html#:~:text=Choose%20%E2%80%9CTools%E2%80%9D%20%3E%20%E2%80%9COrganize,or%20pages%20from%20the%20file.}{Adobe Acrobat} (on all OSs), as well as \href{https://superuser.com/questions/517986/is-it-possible-to-delete-some-pages-of-a-pdf-document}{command line tools}.

\end{document}